\documentclass[a4paper,fleqn]{cas-dc}
\usepackage[utf8]{inputenc}

\usepackage[numbers,sort]{natbib}

\usepackage[english]{babel}
\usepackage[T1]{fontenc}
\usepackage{lmodern}
\usepackage{microtype}
\usepackage{soul}
\usepackage{color}
\usepackage{multirow}
\usepackage{array}
\usepackage{makecell}
\usepackage{graphicx}
\usepackage{xcolor}
\usepackage{placeins} 
\usepackage{dirtytalk} 

\usepackage{footnote}
\makesavenoteenv{tabular}
\makesavenoteenv{table}
\makesavenoteenv{table*}

\usepackage{tikz}
\usetikzlibrary{arrows, positioning, math, external, calc, shapes}

\usepackage[linesnumbered,ruled,vlined,algo2e,resetcount]{algorithm2e}

\begin{document}
\let\WriteBookmarks\relax
\def\floatpagepagefraction{1}
\def\textpagefraction{.001}
\shorttitle{A Survey of Behavior Trees in Robotics and AI}
\shortauthors{Iovino et~al.}

\title [mode = title]{A Survey of Behavior Trees in Robotics and AI}  

\author[1,2]{Matteo Iovino} \ead{matteo.iovino@se.abb.com}
\author[1,3]{Edvards Scukins} \ead{scukins@kth.se}
\author[1,4]{Jonathan Styrud} \ead{jonathan.styrud@se.abb.com}
\author[1]{Petter Ögren} \ead{petter@kth.se}
\author[1]{Christian Smith} \ead{ccs@kth.se}

\address[1]{Division of Robotics, Perception and Learning, Royal Institute of Technology (KTH), Sweden}
\address[2]{ABB Future Labs, hosted at ABB Corporate Research, AI Lab, Sweden}
\address[3]{SAAB Aeronautics, Sweden}
\address[4]{ABB Robotics, Sweden}

\begin{abstract}
Behavior Trees (BTs) were invented as a tool to enable modular AI in computer games, but have received an increasing amount of attention in the robotics community in the last decade. 
With rising demands on agent AI complexity, game programmers found that the Finite State Machines (FSM) that they used scaled poorly and were difficult to extend, adapt and reuse.

In BTs, the state transition logic is not dispersed across the individual states, but organized in a hierarchical tree structure, with the states as leaves. This has a significant effect on \emph{modularity}, which in turn simplifies both synthesis and analysis by humans and algorithms alike. These advantages are needed not only in game AI design, but also in robotics, as is evident from the research being done.

In this paper we present a comprehensive survey of the topic of BTs in Artificial Intelligence and Robotic applications.
The existing literature is described and categorized based on methods, application areas and contributions, and the paper is concluded with a list of open research challenges.
\end{abstract}


\maketitle

\section{Introduction}

\begin{figure}
  \centering
    \includegraphics[width=0.45\textwidth]{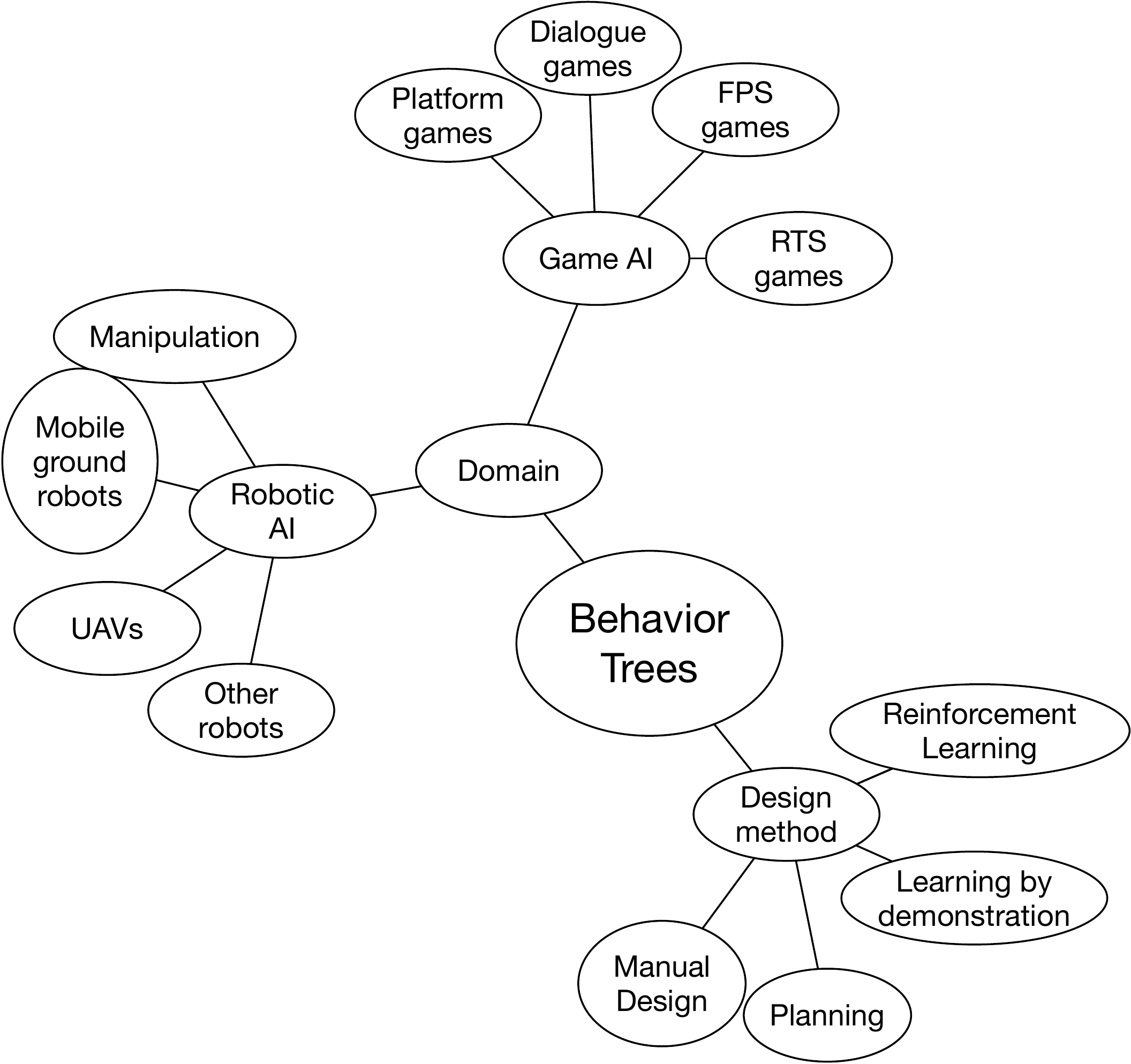}
      \caption{Overview of the topics covered in this survey.}
      \label{overview}
\end{figure}

\begin{figure}
  \centering
    \includegraphics[width=0.48\textwidth]{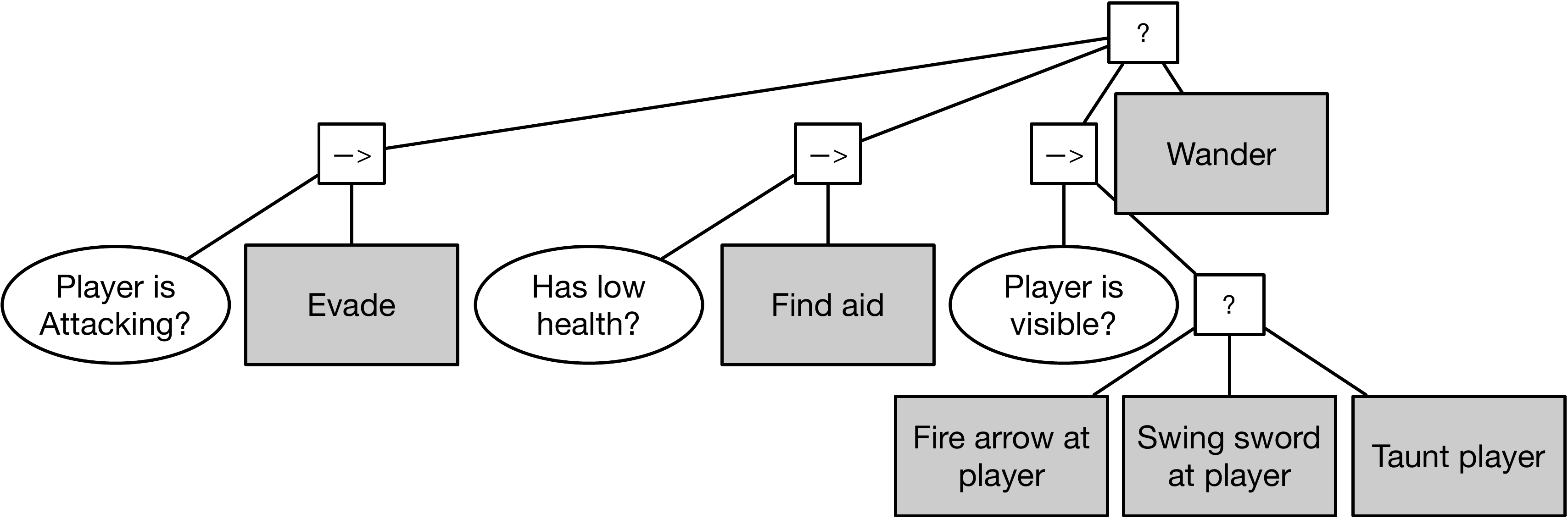}
      \caption{ An example of a BT for a \emph{First Person Shooter} (FPS) game, adapted from \citep{marcotte_behavior_2017}.}
      \label{marcotte17}
\end{figure}

In this paper we survey the area of Behavior Trees (BTs) in AI and robotics applications.\footnote{Note that there is a different concept with the same name, used to handle functional requirements of a system, see e.g. \citep{dromey_requirements_2003}, which will not be addressed here. The term is also used in \citep{iske2001methodology} to denote a general hierarchical structure of behaviors, different from the topic of this paper.} A BT describes a policy, or controller, of an agent such as a robot or a virtual game character. Formally, the policy or controller maps a state to an action. 
We illustrate the idea with a simple example of a BT, see Figure~\ref{marcotte17}. The actions are illustrated by grey boxes ("Evade", "Find aid", "Fire arrow at player", "Swing sword at player", "Taunt player", and "Wander"), while the state of the world is analyzed in the conditions, illustrated by white ovals ("Player is Attacking?", "Has low health?", and "Player is visible?").

We begin by listing a number of \emph{key ideas} behind the design of BTs, while a detailed description of the execution is presented in Section~\ref{sec:classicalBT}.

\begin{itemize}
    \item There is explicit support for the idea of \emph{task hierarchies}, where one task is made up of a number of subtasks, which in turn have subtasks. In Figure~\ref{marcotte17}, \emph{Fighting} is done either with a bow or a sword, and fighting with a bow in turn could include grasping an arrow, placing it on the string, pulling the string, and then releasing it.
    
    \item There is explicit support for \emph{Sequences}, sets of tasks where each task is dependent on the successful completion of the previous, such as first grasping an arrow and then moving it. Thus, the next item in a \emph{Sequence} is only started upon success of the previous one.
    
    \item There is explicit support for \emph{Fallbacks}, sets of tasks where each task represents a different way of achieving the same goal, such as fighting with either a bow or a sword. Thus, the next item in a \emph{Fallback} is only started upon \emph{Failure} of the previous one.
    
    \item There is support for \emph{reactivity}, where \emph{more important tasks interrupt less important ones}, such as disengaging from a fight and finding first aid when health is low.    
      
    \item Finally, \emph{modularity} is improved by the fact that each task, on each level in the hierarchy, has the same interface. It returns either \emph{success}, \emph{failure}, or \emph{running}, which hides a lot of implementation details, while often being enough to decide what to do next.
\end{itemize}

All items above are different aspects of the rules for switching from one task to another. By encoding this task switching logic in the hierarchical structure of the non-leaves of a tree, having basic actions and conditions at the leaves, a surprisingly modular and flexible structure is created. 

Below we first give a brief description of the history of BTs, and then a more detailed description of their core workings.

\subsection{A Brief History of BTs}
\label{subsec:brief_history}
BTs were first conceived by programmers of computer games. Since important ideas were shared on only partially documented blog posts and conference presentations, it is somewhat unclear who first proposed the key ideas, but important milestones were definitely passed through the work of Michael Mateas and Andrew Stern \citep{mateas_behavior_2002}, and Damian Isla \citep{isla_handling_2005}. The ideas were then spread and refined in the community over a number of years, with the first journal paper on BTs appearing in \citep{florez-puga_query-enabled_2009}.
The transition from Game AI into robotics was even later, and 
 independently described in \citep{ogren_increasing_2012} and \citep{bagnell_integrated_2012}.

\subsection{Formulation of the core parts of a BT}
\label{sec:classicalBT}

A BT is a directed tree where we apply the standard meanings of \emph{root}, \emph{child}, \emph{parent}, and \emph{leaf} nodes.
The leaf nodes are called \emph{execution nodes} and the non-leaf
 nodes  are called \emph{control flow nodes}. Graphically, the BT is either drawn with the root to the far left and children to the right, or with the root on top and children below. We will use the latter convention, see the example in Figure~\ref{marcotte17}.

The execution of a BT starts from the root node, that generates signals called \emph{Ticks} with a given frequency.  These signals enable the execution of a node and are then propagated to one or several of the children of the ticked node. A node is executed if, and only if, it receives \emph{Ticks}. The child immediately returns \emph{Running} to the parent,   if its execution is under way, \emph{Success} if it has achieved its goal, or \emph{Failure} otherwise. 

In the classical formulation, there exist three main categories of control flow nodes (\emph{Sequence}, \emph{Fallback} and \emph{Parallel}) and two categories of execution nodes (\emph{Action} and \emph{Condition}), see Table~\ref{core:nodeTable}.

\begin{table*}

\caption{The five node types of a BT.}
\begin{center}
\begin{tabular}{|c|c|c|c|c|c|}
\hline
 \bf{Node type} & {\bf Symbol} & \bf{Succeeds} & \bf{Fails} & \bf{Running} \cr
 
\hline
Sequence & $\rightarrow$ & If all children succeed & If one child fails &If one child returns running \cr

\hline
 Fallback & $?$ & If one child succeeds & If all children fail &If one child returns running \cr

\hline
Parallel & $\rightrightarrows$ & If $\geq M$ children succeed & If $>N-M$ children fail &else \cr
\hline

Action & shaded box& Upon completion & When impossible to complete & During completion \cr
\hline
Condition & white oval & If true & If false & Never  \cr
 \hline
\end{tabular}
\end{center}
\label{core:nodeTable}
\end{table*}%

{\bf Sequences} are used when some actions, or condition checks, are meant to be carried out in sequence, and when the \emph{Success} of one action is needed for the execution of the next.
The Sequence node routes the \emph{Ticks} to its children from the left until it finds a child that returns either \emph{Failure} or \emph{Running}, then it returns \emph{Failure} or \emph{Running} accordingly to its own parent, see Algorithm~\ref{alg:sequence}. It returns \emph{Success} if and only if all its children return \emph{Success}. Note that when a child returns \emph{Running} or \emph{Failure}, the Sequence node does not route the \emph{Ticks} to the next child (if any).
In the case of \emph{Running}, the child is allowed to control the robot, whereas in the case of \emph{Failure}, a completely different action might be executed, or no action at all in the case where the entire BT returns \emph{Failure}.
The symbol of the Sequence node is a box containing the label \say{$\rightarrow$}.

\begin{algorithm2e}[ht]
  \For{$i \gets 1$ \KwSty{to} $N$}
  {
    \ArgSty{childStatus} $\gets$ \FuncSty{Tick(\ArgSty{child($i$)})}\\
    \uIf{\ArgSty{childStatus} $=$ \ArgSty{running}}
    {
      \Return{running}
    }
    \ElseIf{\ArgSty{childStatus} $=$ \ArgSty{failure}}
    {
      \Return{failure}
    }
  }
  \Return{success}
  \caption{Pseudocode of a Sequence node with $N$ children}
  \label{alg:sequence}
\end{algorithm2e}

{\bf Fallbacks}\footnote{Fallback nodes are sometimes also referred to as \emph{selector} or \emph{priority selector} nodes.} are used when a set of actions represent alternative ways of achieving a similar goal. Thus, as shown in Algorithm~\ref{alg:fallback}, the Fallback node
routes the \emph{Ticks} to its children from the left until it finds a child that returns either \emph{Success} or \emph{Running}, then it returns \emph{Success} or \emph{Running} accordingly to its own parent. It returns \emph{Failure} if and only if all its children return \emph{Failure}. Note that when a child returns \emph{Running} or \emph{Success}, the Fallback node does not route the \emph{Ticks} to the next child (if any).
The symbol of the Fallback node is a box containing the label \say{$?$}.

\begin{algorithm2e}[t]
  \For{$i \gets 1$ \KwSty{to} $N$}
  {
    \ArgSty{childStatus} $\gets$ \FuncSty{Tick(\ArgSty{child($i$)})}\\
    \uIf{\ArgSty{childStatus} $=$ \ArgSty{running}}
    {
      \Return{running}
    }
    \ElseIf{\ArgSty{childStatus} $=$ \ArgSty{success}}
    {
      \Return{success}
    }
  }
  \Return{failure}
  \caption{Pseudocode of a Fallback node with $N$ children}
  \label{alg:fallback}
\end{algorithm2e}

{\bf Parallel} nodes tick all the children simultaneously. Then, as shown in Algorithm~\ref{alg:parallell}, if $M$ out of the $N$ children return \emph{Success}, then so does the parallel node. If more than $N-M$ return \emph{Failure}, thus rendering  success impossible, it returns \emph{Failure}. If none of the conditions above are met, it returns running. The symbol of the Parallel node is a box containing the label \say{$\rightrightarrows$}.
\begin{algorithm2e}[t]
  \For{$i \gets 1$ \KwSty{to} $N$}
  {
    \ArgSty{childStatus}(i) $\gets$ \FuncSty{Tick(\ArgSty{child($i$)})}\\
    \uIf{$\Sigma_{i: \ArgSty{childStatus}(i)=success}1\geq M$}
    {
      \Return{Success}
    }
    \ElseIf{$\Sigma_{i: \ArgSty{childStatus}(i)=failure}1 > N-M$}
    {
      \Return{failure}
    }
  }
  \Return{running}
  \caption{Pseudocode of a parallel node with $N$ children and success threshold $M$}
   \label{alg:parallell}
\end{algorithm2e}

{\bf Action} nodes  typically execute a command when receiving \emph{Ticks}, such as e.g. moving the agent. If the action is successfully completed, it returns \emph{Success}, and  if the action has failed, it returns \emph{Failure}. While the action is ongoing it returns \emph{Running}. Actions are represented by shaded boxes.

{\bf Condition} nodes check a proposition upon receiving Ticks. It returns \emph{Success} or \emph{Failure} depending on if the proposition holds or not. Note that a Condition node never returns a status of \emph{Running}. Conditions are thus technically a subset of the Actions, but are given a separate category and graphical symbol to improve readability of the BT and emphasize the fact that they never return running and do not change the world or any internal states/variables of the BT. Conditions are represented by white ovals.

\subsection{Overview}

The purpose of this paper is to provide an overview of the current body of literature covering BTs in robotics. To do this, the Google Scholar~\footnote{scholar.google.com}, Scopus~\footnote{www.scopus.com}, and Clarivate Web of Science~\footnote{www.webofknowledge.com} databases were searched with the keyword "Behavior Tree" and alternative spelling "Behaviour Tree". Only papers written in English have been taken into account. This resulted in 297 papers. Removing the papers referring to Behavior Trees as functional requirements (as in e.g.~\citep{dromey_requirements_2003}) and cleaning the list from dead links and duplicates, the final number as of April 24, 2020 was 166 papers. All papers are classified according to topic matter and application areas in Table~\ref{tab:taxonomy}, and this structure is also used throughout the paper. 

The remainder of the paper is organized as follows. In Section \ref{sec:theory} we review the papers that analyze the core theory of BTs. Then, in Section \ref{sec:appl}, we review the work done in the different areas of application. The different methods being used are then analyzed in Section \ref{sec:method}. Finally, some common libraries for implementing BT are discussed in Section \ref{sec:impl},  some open challenges are listed in Section \ref{sec:end}, and the conclusions are drawn in Section \ref{sec:conclusions}.

\begin{table*}
\small
\centering

\caption{Taxonomy of Behavior Trees papers}

\begin{tabular}{c p{0.4cm} p{2.99cm}|p{1.40cm}|p{1.4cm}|p{1.9cm}|p{1.40cm}|p{1.40cm}|}

& & \multicolumn{6}{c}{Applications} \vspace{4pt}\\

\cline{3-8}
& & \multicolumn{1}{|c|}{\multirow{2}{*}{Game AI}} & \multirow{2}{*}{\parbox[c]{1.8cm}{Combat\\Training\\Simulation}} & \multicolumn{3}{c|}{Robotics} & \multirow{2}{*}{Others}\\

\cline{5-7}
\multicolumn{2}{c}{Methodology} & \multicolumn{1}{|c|}{} & & Manipulation & \parbox[c]{1.4cm}{Mobile Ground\\Robots} & \parbox[c]{1.4cm}{Aerial and \\Underwater \\Vehicles} & \\

\hline
\multicolumn{1}{|c|}{\multirow{2}{*}{Learning}} & \multicolumn{1}{c|}{RL/GP} 
&
\citep{colledanchise_behavior_2018}
\citep{zhang_behavior_2018}
\citep{nicolau_evolutionary_2017}
\citep{hutchison_evolving_2010}
\citep{perez_evolving_2011}
\citep{hoff_evolving_2016}
\citep{pena_learning_2012}
\citep{zhang_learning_2018}
\citep{colledanchise_learning_2019}
\citep{oakes_practical_2013}
\citep{dey_ql-bt_2013}
\citep{estgren_behaviour_2017}
\citep{paduraru_automatic_2019}
\citep{schwab_capturing_2015}
\citep{fu_reinforcement_2016}
\citep{lim_.i._2009}
\citep{kartasev_integrating_2019}
\citep{zhu_behavior_2019}
\citep{hallawa_evolving_2020}
&
\citep{yao_adaptive_2015}
\citep{berthling-hansen_automating_2018}
\citep{eilert_learning_2019}
\citep{yao_adaptive_2015-1}
\citep{zhang_combining_2017}
\citep{yao_tactics_2017}
\citep{zhang_modeling_2017}
\citep{zhang_towards_2016}
& & 
\citep{jones_evolving_2018}
\citep{jones_two_2018}
\citep{neupane_emergence_2019}
\citep{neupane_designing_2019}
\citep{banerjee_autonomous_2018}
&
\citep{scheper_behavior_2015}
&
\citep{sprague_adding_2018} 
\citep{neupane_learning_2019}
\\

\cline{2-8}
\multicolumn{1}{|c|}{} & \multicolumn{1}{c|}{By Demo} & 
\citep{pereira_framework_2015} \newline
\citep{robertson_building_2015} \newline
\citep{palma_extending_2011} \newline
\citep{sagredo-olivenza_trained_2019} \newline
\citep{sagredo-olivenza_combining_2017}
\citep{palma_combining_2011}
\citep{florez-puga_dynamic_2008}
\citep{buche_orion_2020}
& & 
\citep{french_learning_2019}

& 
\citep{french_learning_2019}
& &
\\

\hline
\multicolumn{2}{|r|}{Hand Coded} & 
\citep{isla_handling_2005}
\citep{cutumisu_architecture_2009}
\citep{millington_artificial_2009}
\citep{florez-puga_query-enabled_2009}
\citep{shoulson_parameterizing_2011}
\citep{oshea_extending_2011} 
\citep{becroft_aipaint_2011}
\citep{johansson_emotional_2012}
\citep{ocio_adapting_2012}
\citep{merrill_building_2013}
\citep{shoulson_event-centric_2013}
\citep{colledanchise_performance_2014}
\citep{colledanchise_how_2014}
\citep{plch_ai_2014}
\citep{pereira_framework_2015}
\citep{dominguez_automated_2015}
\citep{subagyo_simulation_2016}
\citep{li_game_2016}
\citep{waltham_analysis_2016}
\citep{marcotte_behavior_2017}
\citep{rabin_game_2017}
\citep{dagerman_high-level_2017}
\citep{agis_event-driven_2020}
\citep{sun_animating_2012}
\citep{geraci_authoring_2015}
\citep{li_baap_2010}
\citep{ripamonti_believable_2017}
\citep{johansson_comparing_2012}
\citep{kapadia_computer-assisted_2015}
\citep{florez-puga_empowering_2011}
\citep{triebel_generation_2012}
\citep{othman_implementing_2014}
\citep{wang_object_2017}
\citep{delmer_behavior_2012}
\citep{martens_villanelle_2019}
\citep{buche_autonomous_2018}
\citep{sagredo-olivenza_supporting_2015}
\citep{gonzalez_calero_artificial_2011}
\citep{tomai_simulating_2013}
\citep{brown_facilitating_2012}
\citep{ocio_building_2015}

& 
\citep{gong_agent_2019}
\citep{xiaobing_visual_2010}
\citep{geng_hybrid_2011}
\citep{woolley_cognitive_2016}
\citep{kamrud_unified_2017}
& 
\citep{marzinotto_towards_2014} 
\citep{marzinotto_flexible_2017}
\citep{bagnell_integrated_2012} 
\citep{rovida_motion_2018}
\citep{guerin_framework_2015}
\citep{paxton_costar_2017}
\citep{paxton_user_2017}
\citep{paxton_evaluating_2018}
\citep{colledanchise_performance_2014}
\citep{colledanchise_how_2017}
\citep{colledanchise_analysis_2019}
\citep{csiszar_behavior_2017}
&
\citep{olsson_behavior_2016} 
\citep{colledanchise_advantages_2016}
\citep{colledanchise_how_2014}
\citep{colledanchise_improving_2018}
\citep{colledanchise_analysis_2019}
\citep{jiang_laair_2018}
\citep{kim_architecture_2018}
\citep{hannaford_simulation_2016}
\citep{marzinotto_towards_2014}
\citep{abiyev_control_2013}
\citep{abiyev_robot_2016}
\citep{siqueira_context-aware_2015}
\citep{siqueira_semantic_2016}
\citep{yang_hierarchical_2019}
\citep{giunchiglia_conditional_2019}
\citep{coronado_robots_2019}
\citep{macenski_marathon_2020}
& 
\citep{colledanchise_advantages_2016}
\citep{colledanchise_how_2014}
\citep{klockner_behavior_2013}
\citep{klockner_behavior_2015}
\citep{sprague_improving_2018}
\citep{ogren_increasing_2012}
\citep{klockner_interfacing_2018}
\citep{klockner_modelica_2014}
\citep{castano_safe_2019}
\citep{lan_modular_2018}
\citep{golluecke_behavior_2018}
\citep{safronov_asynchronous_2019}
\citep{goudarzi_mission_2019}

&
\citep{sprague_adding_2018}
\citep{deneke_conceptual_2017}
\citep{hu_semi-autonomous_2015}
\citep{hannaford_behavior_2018}

\\

\hline
\multicolumn{2}{|r|}{Planned Analytically} &
\citep{colledanchise_synthesis_2017}
\citep{holzl_reasoning_2015} & &
\citep{colledanchise_towards_2019}
\citep{tumova_maximally_2014}
\citep{rovida_extended_2017}
\citep{marzinotto_flexible_2017} 
\citep{paxton19}
& 
\citep{colledanchise_towards_2019}
\citep{rovida_extended_2017}
\citep{kuckling_behavior_2018}
\citep{tumova_maximally_2014}
\citep{segura-muros_integration_2017}
\citep{zhou_autonomous_2019} &
\citep{tadewos_--fly_2019}
&
\citep{tadewos_automatic_2019}
\\

\hline
\multicolumn{2}{|r|}{Others} &
\citep{llanso_self-validated_2009}
\citep{sekhavat_behavior_2017}
\citep{plch_modular_2014}
\citep{ji_research_2014}
\citep{tremblay_understanding_2012}
\citep{balint_understanding_2018}
\citep{zhang_coupling_2019}
\citep{belle_programming_2019}

\citep{mateas_behavior_2002}
\citep{weber_reactive_2010}
\citep{weber_building_2011}
\citep{simpkins_towards_2008}
& 
\citep{ramirez_integrated_2018}
&
\citep{berenz_playful_2018} &
\citep{berenz_playful_2018} & & 
\citep{safronov_node_2020}
\\

\hline
\end{tabular}
\label{tab:taxonomy}
\end{table*}

\section{Fundamental theory}
\label{sec:theory}

\begin{table}
    \centering
    \caption{Areas of fundamental theory for BTs}
\begin{center}
    \begin{tabular}{| l | p{3.5cm} |}
    \hline
     Fundamental Theory Area & Papers  \\ \hline
    Establishing the BT core & 
    \citep{mateas_behavior_2002} 
    \citep{isla_handling_2005} 
    \citep{florez-puga_query-enabled_2009} 
    \citep{rabin_behavior_2013}\\
    Architecture comparisons & 
    \citep{isla_handling_2005} 
    \citep{ogren_increasing_2012}
    \citep{marzinotto_towards_2014}
    \citep{colledanchise_how_2017}
    \citep{colledanchise_how_2016}
    \citep{chen_development_2018} \\
    Variations on Fallback node & 
    \citep{merrill_building_2013}
    \citep{hannaford_simulation_2016}\\
    Convergence analysis  & 
    \citep{colledanchise_how_2016}
    \citep{,colledanchise_how_2017}
    \citep{rovida_extended_2017}
    \citep{sprague_improving_2018}
    \citep{paxton19}\\
    Stochastic analysis  & \citep{colledanchise_performance_2014} \citep{hannaford_simulation_2016} \citep{hannaford_hidden_2019}\\
    Analysis of Parallel node  &
    \citep{csiszar_behavior_2017}
    \citep{colledanchise_improving_2018} \citep{colledanchise_analysis_2019} \citep{rovida_motion_2018} \\
    Parameters and data passing  &\citep{shoulson_parameterizing_2011} \\
    Multi agent BTs  & \citep{colledanchise_advantages_2016}\\
    Learning BTs  & See Section \ref{sec:learning}\\ 
    Planning BTs  & See Section \ref{sec:planning}\\
    \hline
    \end{tabular}
\end{center}
\label{tab:theory}
\end{table}{}

In this section we describe the theoretical evolution of the BT core concept, including some extensions and tools for theoretical analysis. We will roughly follow the structure provided by Table~\ref{tab:theory}.

Sequences were introduced in \citep{mateas_behavior_2002}, as composed of behaviors that can either \emph{succeed} or \emph{fail}. There was no notion of returning \emph{running} in this early version, an important concept that is needed to provide \emph{reactivity}, as will be described below.

Fallbacks were also present in an early form in \citep{mateas_behavior_2002}, where it is described how actions to achieve a given result are collected, and their corresponding preconditions are checked. If more than one are satisfied, the one with the highest specificity is chosen, see Figure~\ref{mateas02}. Fallbacks were subsequently given a more refined form in terms of the prioritized lists in \citep{isla_handling_2005} and \citep{florez-puga_query-enabled_2009}.

\emph{Reactivity}, the ability to stop an action before it completes (either \emph{succeeds} or \emph{fails}) and switch to a more urgent action, is a very important quality of an agent. In modern BTs this is achieved using the \emph{running} return status, in combination with recurrent ticks from the root, prompting a reevaluation of conditions and actions, see Algorithms~\ref{alg:sequence}-\ref{alg:fallback}. It was suggested in \citep{florez-puga_query-enabled_2009} that ``conditions are reevaluated \mbox{after} a given number of game ticks (frames), when certain game events occur or when the active behavior terminates''. In the formulation described in Section~\ref{sec:classicalBT}, reevaluation takes place when the tick is sent from the root with a given frequency, to provide \emph{reactivity} to external events. 
If external changes can somehow be tracked centrally, the same \emph{reactivity} would be achieved by only ticking ``when certain game events occur''. However, only ticking ``when the active behavior terminates'' is vastly different, as it would imply that the agent always finishes its current task before starting a new one, effectively preventing any reactions to unexpected high priority events such as 
fire alarms, safety margin infractions, or the appearance of other threats or opportunities. In fact, the absence of a reactive capability was pointed out as the main drawback of BTs in \citep{millington_artificial_2009}, where a BT formulation without \emph{running} was used. This is  somewhat ironic since today \emph{reactivity}, together with \emph{modularity}, is often listed as some of the main advantages of using BTs.
This use of \emph{Ticks}, and the \emph{running} return status as described in Algorithms~\ref{alg:sequence}-\ref{alg:fallback}, including the Parallel node of Algorithm~\ref{alg:parallell}, was described in \citep{rabin_behavior_2013}.

Prior to BTs, the Finite State Machine (FSM) has long been used as a way of organizing task switching, and a BT can be seen as a FSM with special structure, as noted in \citep{isla_handling_2005}. The relationship between FSMs and BTs was further explored in \citep{ogren_increasing_2012,marzinotto_towards_2014,colledanchise_how_2017,chen_development_2018}. In these papers, it was shown how to create a BT that works like an FSM, by keeping track of the current state as an external variable on a \emph{blackboard}, and an FSM that works like a BT by letting all ticks and return statuses correspond to state transitions. Thus, the two structures are equivalent in terms of what overall behavior can be created, much in the same way as an algorithm can be implemented in any general programming language. However, from a practical standpoint, the differences are often significant.

FSMs use state transitions, that transfer the control of the agent in a way that is very similar to the GOTO statement, that is now abolished in high level programming languages, \citep{dijkstra1968go}. 
In contrast, the control transfer in a BT resembles function calls, with subtrees being ticked, and returning \emph{success}, \emph{failure} or \emph{running}. Furthermore, assuming $n$ possible actions (represented as states), the $n^2$ possible transitions of an FSM rapidly turns to a single large monolithic structure that is very complex to debug, update and extend \citep{oshea_extending_2011}. For a BT on the other hand, each subtree is a natural level of abstraction, with all subtrees sharing the same interface of incoming ticks and outgoing return statuses. 

Beyond the FSMs, the relationship between other switching architectures and BTs have been explored. It was shown in \citep{colledanchise_how_2017} how Decision Trees and the Subsumption Architecture are generalized by BTs and in \citep{colledanchise_how_2016} how the Teleo-Reactive approach \citep{nilsson1993teleo} and AndOrTrees are generalized.  

The standard Fallback node in Algorithm~\ref{alg:fallback} has a static priority order of all its children, and it has been noted by several authors that there are many cases where it makes sense to update this order based on the state of the world. One such approach, Utility BTs, was suggested in \citep{merrill_building_2013}. Here each subtree of the BT is able to return a utility value, and there are special \emph{Utility Fallbacks}, that sort their children based on this utility score. Leaf nodes then have to implement such a utility estimate, whereas other interior nodes have to be able to aggregate utility based on their children. This can be done in many ways, and the one suggested in \citep{merrill_building_2013} is to report the highest child utility as the utility of both Fallbacks and Sequences.

Another way of addressing static Fallbacks, based on estimated success probabilities, was suggested in \citep{hannaford_simulation_2016}. Here the system gathers data on the \emph{success}/\emph{failure} outcomes of all leaves. Thereby it is able to estimate the success probability of each leaf, and reorder the children of a Fallback accordingly. This result is data driven and may therefore also be considered to be a learning approach. Such approaches are described in detail in Section~\ref{sec:learning}.

Given such statistics on \emph{success} and \emph{failure} probabilities of leaves, it is natural to expand to \emph{success} and \emph{failure} probabilities of entire subtrees, as in \citep{colledanchise_performance_2014}. In addition, if given estimates of execution time of the leaf nodes, in terms of probability density functions over time for success and failure, these estimates can be aggregated upwards and render execution time estimates for all subtrees.
Finally, the stochastic analysis of BTs are taken one step further in \citep{hannaford_hidden_2019}, where Hidden Markov Models (HMM) are connected to BTs where only noisy observations are available from the execution. Given observation data, it is shown how HMM tools can be applied to estimate state transition probabilities of the BT, to estimate what transitions are the most likely given some data, and how likely the output of some data is, given a set of parameters.

The  Sequence node in Algorithm~\ref{alg:sequence} is reactive in the sense that it constantly reevaluates its children. This can sometimes cause problems when executing actions that do not leave a trace of their successful completion. An example of this was described in  \citep{klockner_behavior_2015} with an agent following a waypoint path. After passing a waypoint the agent switches to the next. However, if the success condition is the distance to the waypoint, the first action will not return \emph{Success} when leaving the first waypoint on its way to the second. One common solution to this problem is to create a new node type, which is a Sequence node with a memory\footnote{sometime called Sequence*} of what child is active, and skip over a child that has already returned \emph{Success} at some earlier point in time. However, as noted in  \citep{klockner_behavior_2015} the same result can be achieved by adding a \emph{decorator} node above each waypoint action that keeps returning \emph{success} without ticking its child, after the initial \emph{success} has been returned.

A reasonable question to ask is in what parts of the state space a BT will return \emph{success} or \emph{failure}. This problem was addressed for Teleo-reactive designs in~\citep{nilsson1993teleo} and carried over to BTs in \citep{colledanchise_how_2017} and~\citep{rovida_motion_2018}. In order to do a formal statespace analysis, a functional model was proposed in~\citep{colledanchise_how_2014,colledanchise_how_2017}. Using tools from control theory, such as \emph{region of attraction}, and \emph{exponential stability}, properties such as robustness (regions of attraction), safety (avoidance of some regions) and efficiency (convergence within upper time bounds) were addressed. By showing how the properties carry over across sequence and fallback compositions, the analysis can be done for larger BTs.

The analysis of robustness, safety and efficiency was continued in~\citep{sprague_adding_2018}, where it was shown how a learning subtree, with possibly unreliable performance, could be added while some guarantees of robustness, safety and efficiency could still be given for the overall BT.

In \citep{paxton19}, a concept that is very similar to BTs was introduced in the form of Robust Logical Dynamical Systems. Convergence of Sequences of such components was proven, together with a stochastic analysis of convergence times when faced with a bounded number of external state transitions.

\citep{rovida_extended_2017} extend the core notation of BTs by adding pre- and post-conditions to the Action nodes and thus removing Condition nodes. This modified model is called \emph{extended Behavior Tree} (eBT) and is meant to be interfaced with Hierarchical Task Network (HTN) planning. This planning algorithm is also responsible for rearranging the tree, optimizing the execution time and the resources used, but without compromising the correctness of the tree. The authors also state the differences between their model and the standard one. This framework is also used in~\citep{rovida_motion_2018}, where it is combined with a \emph{Motion Generator} (MG), responsible for superposing individual and independent motion primitives to generate a hybrid motion. This setup allows the activation of different primitives concurrently, through the Parallel operator, hence crossing the limits stated also in~\citep{colledanchise_improving_2018}.

Most actions of a BT require data on the world, such as positions of objects and other agents, or the amount of charge left in a battery.
\citep{shoulson_parameterizing_2011} introduce parametrization in Behavior Trees. Instead of using a \emph{blackboard}, i.e., a repository of data available to all the nodes of the tree as in~\citep{rovida_motion_2018}, they propose to add parameters to a node such that the node stores all the parameters needed for its subtree execution. This is done in three different ways: it can be hard coded, taken from the characteristics of the world and the agents, or satisfied with \emph{Parametrized Action Representation} (PAR) arguments, which allow for reuse and encapsulation of the sub-tree. 

\citep{colledanchise_improving_2018} introduce the notion of Concurrent Behavior Trees (CBT) as a way of improving the Parallel operator in standard BTs. Indeed, the Parallel operator is normally used when the sub-tree tasks are orthogonal, i.e., independent. However, this is not the case for most robotic applications, and the use of the Parallel operator can help mitigating the curse of dimensionality. The standard Parallel operator inherits the well known problems of parallel computing, namely  \emph{process synchronization} (when one or more processes are queued and the latter ones have to wait for the end of the execution of the former) and the \emph{data synchronization} (when regulations is needed on data access). Therefore, the CBT is equipped with a \emph{Progress Function}, which indicates the state of the BT execution, and a \emph{Resource Function}, which lists a set of resources needed for the execution at each state. Finally, two new Parallel operators are defined. The \emph{Synchronized Parallel operator}, stipulates that all the children share a minimum progress value that they have to reach in order to proceed with the execution, hence waiting for the slowest one. The \emph{Mutually Exclusive Parallel operator} only executes a child if it uses resources which are not needed by other running children. These new operators were tested on a mobile robot.
The parallel node was also discussed and expanded by~\citep{csiszar_behavior_2017} in combination with event driven nodes, in the context of an architecture for industrial robot control based on BTs.

\section{Applications}
\label{sec:appl}

In this section we describe different application areas, see Figure~\ref{overview}, and give a short overview of how BTs have been used in the different areas.

\subsection{Game AI and Chatbots}

\begin{figure}
  \centering
    \includegraphics[width=0.38\textwidth]{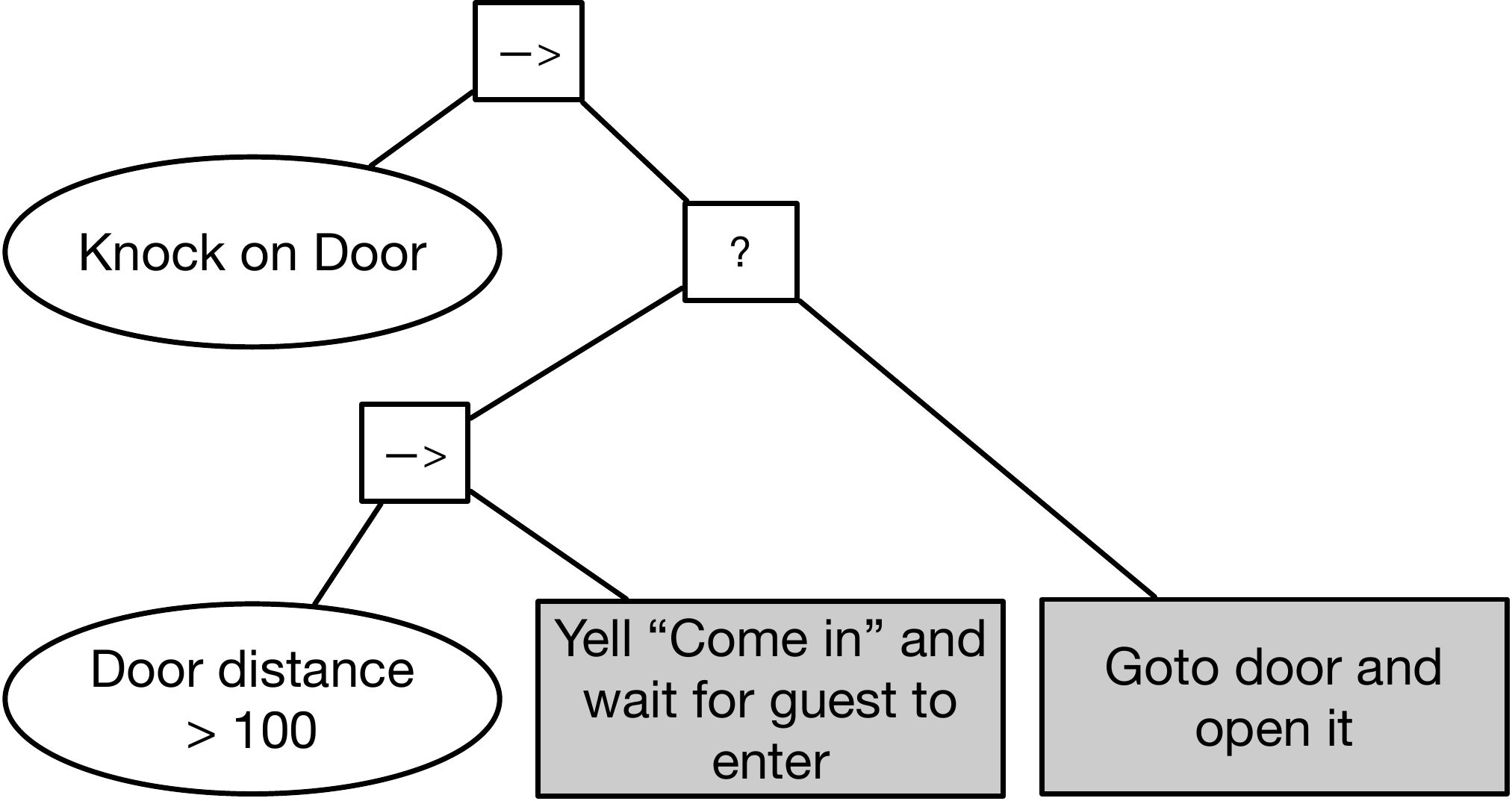}
      \caption{ A BT from a dialogue game, describing how to react when someone knocks on the door, adapted from \citep{mateas_behavior_2002}. }
      \label{mateas02}
\end{figure}

As described in Section~\ref{subsec:brief_history}, BTs were first created for a dialogue game AI in \citep{mateas_behavior_2002}, see Figure~\ref{mateas02}, and have since become a standard tool for developing the AI of non-player characters in many different game genres, 
 from strategy games to shooters. 
 
Below we discuss the use of BTs in these different game categories in more detail.

\begin{table}[]
  \caption{BTs for Game AI}
    \label{tab:gaming}
\begin{center}
    \begin{tabular}{| p{2.5cm} | p{4cm} |}
    \hline
     Type of Game & Papers  \\ \hline
    Real time strategy games (RTS) & \citep{delmer_behavior_2012}, \citep{robertson_building_2015}, \citep{weber_building_2011}, \citep{hutchison_evolving_2010}, \citep{hoff_evolving_2016}, \citep{palma_extending_2011}, \citep{oakes_practical_2013}, \citep{weber_reactive_2010}, \citep{palma_combining_2011}, \citep{lim_.i._2009}\\
    First Person Shooters (FPS) &  \citep{florez-puga_dynamic_2008}, \citep{johansson_emotional_2012}, \citep{isla_handling_2005}, \citep{florez-puga_query-enabled_2009}, \citep{sagredo-olivenza_trained_2019}, \citep{ripamonti_believable_2017}, \citep{buche_orion_2020} \\
    Platform games &  \citep{becroft_aipaint_2011}, \citep{zhang_behavior_2018}, \citep{nicolau_evolutionary_2017}, \citep{perez_evolving_2011}, \citep{dagerman_high-level_2017}, \citep{zhang_learning_2018}, \citep{colledanchise_learning_2019},  \\
    Dialogue games & \citep{mateas_behavior_2002}, \citep{cutumisu_architecture_2009}, \citep{rabin_game_2017}, \citep{dominguez_automated_2015}, \citep{sun_animating_2012}, \citep{kapadia_computer-assisted_2015}, \citep{coronado_development_2018}\\
    \hline
    \end{tabular}
\end{center}
  
\end{table}{}

\subsubsection{Real-time strategy games (RTS)} 
Real-time strategy games require players to make tactical decisions,
such as taking control of some given area, or aim an attack on some particular enemy facilities.
 In such games, there is a large number of different units,  that can be assigned to specific tasks based on their different strengths and weaknesses \citep{palma_combining_2011}.   
These units can also be combined into squads, platoons and so on up to an entire army, and BTs have been used to make decisions on many such levels of abstraction~\citep{delmer_behavior_2012}. 
An early paper on RTS AI is~\citep{hutchison_evolving_2010},
where evolutionary methods were used to create BTs that could defeat a hand coded AI for the RTS game DEFCON.
Other examples of evolutionary methods to create BTs include \citep{oakes_practical_2013}, \citep{hoff_evolving_2016}, where an AI for turn based strategy games was developed.

In \citep{robertson_building_2015}, the authors generated BTs based on recurring actions  executed by human experts. The proposed approach did not need any action model or rewards, but required a significant number of training traces.
An alternative approach using manually injected knowledge was proposed in 
 \citep{palma_extending_2011}.
One of the most well known, and complex, RTS games is StarCraft. In~\citep{weber_reactive_2010}, StarCraft was used to demonstrate the ability of A Behavior Language (ABL) to do reactive planning, and a human-level agent was proposed in~\citep{weber_building_2011}.

\subsubsection{First Person Shooters (FPS)}
One of the first papers on BTs describe how they are used to create the AI of the highly popular FPS game Halo 2~\citep{isla_handling_2005}.
In FPS games, the player interacts with a large number of Non-Player Characters in a highly dynamical combat environment, and an example of the decisions made and actions taken can be seen in Figure~\ref{marcotte17}.

The problem of reusing BTs in situations where the available actions of an agent increases over time was studied in \citep{florez-puga_dynamic_2008}. An approach where actions are categorized is proposed, in which the leaves of the BT make high level queries to find the action suitable for the situation at hand.
This work is then continued in~\citep{florez-puga_query-enabled_2009,gonzalez_calero_artificial_2011}, where a higher abstraction layer is added above the BT. Having a set of BTs for achieving different goals, case based reasoning is used to chose the appropriate BT to invoke.

Additional modifications have been applied to FPS BTs by including time, risk perception, and emotions \citep{johansson_emotional_2012}, or by learning BTs from traces that were obtained during human game play (programming by demonstration)~\citep{sagredo-olivenza_trained_2019,buche_orion_2020}.

\subsubsection{Platform games}

\begin{figure}
  \centering
    \includegraphics[width=0.45\textwidth]{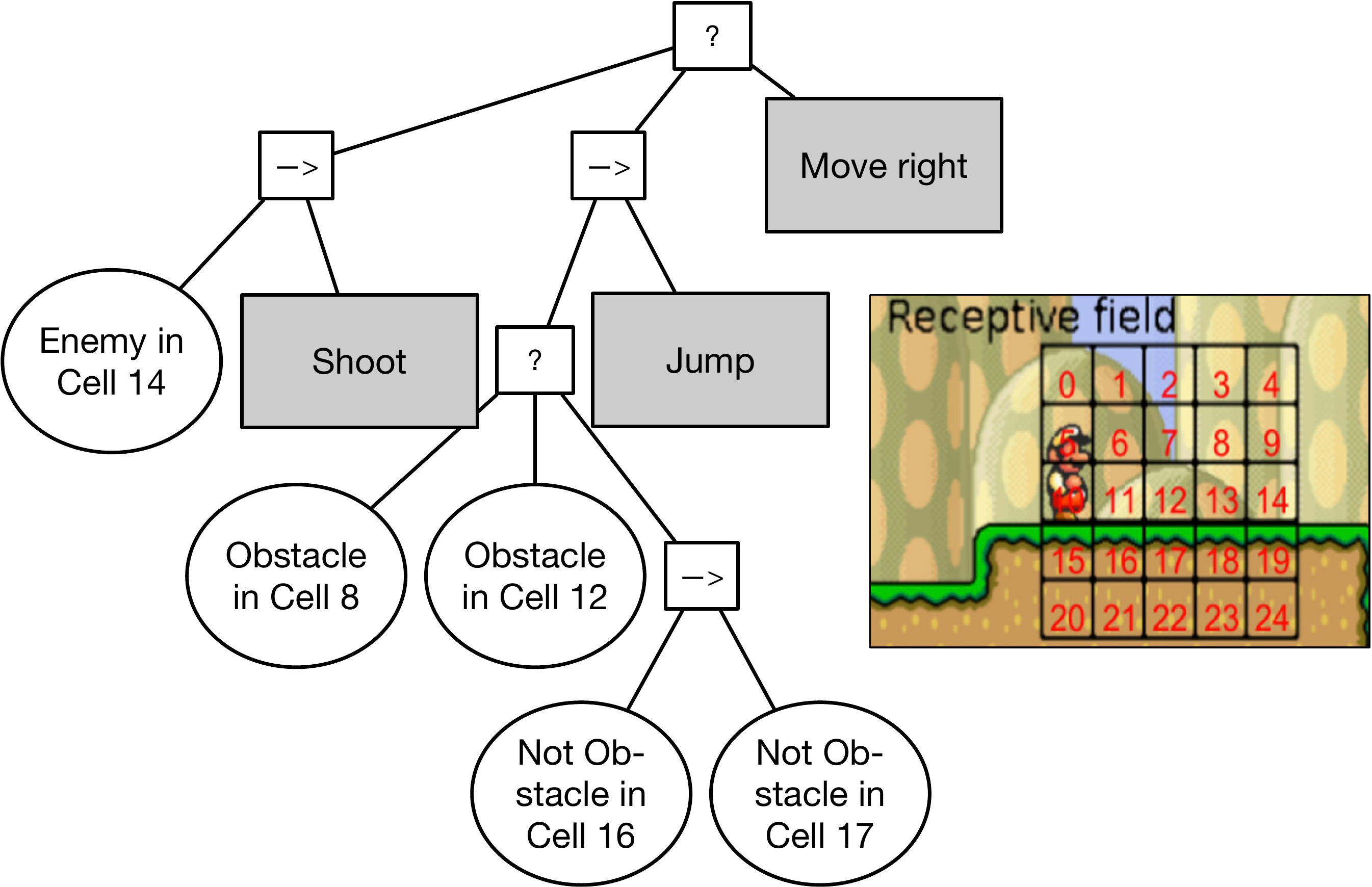}
      \caption{ An example of a BT for a platform game, that was created using an evolutionary learning approach, adapted from \citep{colledanchise_learning_2019}. The cell numbers of the conditions can be found in the illustration to the right.}
      \label{colledanchise19}
\end{figure}

Platform games are  games where the player control a character in a 2D environment and move between a set of platforms~\citep{karakovskiy2012mario}. The Mario AI framework emulates a popular platform game, and has become a widely used tool to test evolutionary approaches to learning a BT~\citep{perez_evolving_2011,zhang_behavior_2018,colledanchise_learning_2019,nicolau_evolutionary_2017}.  Pac-Man is not technically a platform game, but shares the 2D structure, and has been used to test Monte Carlo Tree Search using BTs~\citep{dagerman_high-level_2017}, hybrid evolving BTs~\citep{zhang_learning_2018} and tools with focus on supporting manual BT design~\citep{becroft_aipaint_2011}.

\subsubsection{Dialogue games}

In an effort to make  game environments more realistic, designers sometimes enable non-player characters to interact with the player not only through actions, but also through spoken dialogue.

This was the motivating application for the conception of BTs in~\citep{mateas_behavior_2002}, and has later been explored in~\citep{cutumisu_architecture_2009,sun_animating_2012,kapadia_computer-assisted_2015,coronado_development_2018}. In~\citep{cutumisu_architecture_2009} situations such as a waiter making conversation while retrieving orders is considered, while~\citep{coronado_development_2018} investigates a human-robot interaction scenario.
Interactive narratives are explored in~\citep{kapadia_computer-assisted_2015}, where two virtual teddy bears ask the player to find them a ball to play with, while in~\citep{sun_animating_2012}, conversation types such as buyer and seller negotiations and simple asking-answering are investigated.


\subsection{Robotics}
The transition of BTs from Game AI into robotics was independently proposed and described in 2012, in~\citep{bagnell_integrated_2012,ogren_increasing_2012}. The former applies BTs to  object grasping and dexterous manipulation, and the latter proposes to use BTs for UAV control. In this section we divide the papers into three categories based on the type of robot that BTs are applied to: \emph{manipulators}, \emph{mobile ground robots}, and \emph{aerial and underwater robots}. 
A division into even finer categories can be found in Table~\ref{tab:robotic_systems}.

\begin{table}[]
 \caption{BTs for different kinds of robot systems.}
    \label{tab:robotic_systems}
\begin{center}
    \begin{tabular}{|l|p{4.5cm}|}
    \hline
    Robotic System & Papers  \\ \hline
    Manipulators &
    \citep{marzinotto_flexible_2017},
    \citep{bagnell_integrated_2012},
    \citep{rovida_motion_2018},
    \citep{guerin_framework_2015},
    \citep{paxton_costar_2017},
    \citep{paxton_user_2017},
    \citep{paxton_evaluating_2018},
    \citep{colledanchise_performance_2014},
    \citep{colledanchise_how_2017},
    \citep{csiszar_behavior_2017},
    \citep{colledanchise_towards_2019},
    \citep{berenz_playful_2018}\\
    Mobile Manipulators &
    \citep{colledanchise_advantages_2016},
    \citep{colledanchise_improving_2018},
    \citep{colledanchise_analysis_2019},
    \citep{colledanchise_towards_2019}, 
    \citep{rovida_extended_2017},
    \citep{french_learning_2019},
    \citep{jiang_laair_2018},
    \citep{kim_architecture_2018},
    \citep{segura-muros_integration_2017},
    \citep{giunchiglia_conditional_2019},
    \citep{zhou_autonomous_2019}\\
    Wheeled Robots &
    \citep{siqueira_context-aware_2015},
    \citep{siqueira_semantic_2016},
    \citep{banerjee_autonomous_2018},
    \citep{abiyev_control_2013},
    \citep{abiyev_robot_2016},
    \citep{macenski_marathon_2020}\\
    Robot Swarms &
    \citep{kuckling_behavior_2018},
    \citep{jones_evolving_2018},
    \citep{jones_two_2018},
    \citep{neupane_emergence_2019},
    \citep{neupane_designing_2019},
    \citep{yang_hierarchical_2019}\\
    Humanoid Robots &
    \citep{colledanchise_how_2017},
    \citep{berenz_playful_2018},
    \citep{marzinotto_towards_2014},
    \citep{tumova_maximally_2014},
    \citep{hannaford_simulation_2016},
    \citep{coronado_robots_2019}\\
    Autonomous vehicles &
    \citep{colledanchise_how_2014},
    \citep{olsson_behavior_2016}\\
    Aerial Robots &
    \citep{klockner_behavior_2013},
    \citep{klockner_modelica_2014},
    \citep{klockner_behavior_2015},
    \citep{klockner_interfacing_2018},
    \citep{scheper_behavior_2015},
    \citep{lan_modular_2018},
    \citep{castano_safe_2019},
    \citep{li_uav_2019},
    \citep{lan_autonomous_2019},
    \citep{goudarzi_mission_2019}\\
    Underwater Robots &
    \citep{sprague_improving_2018}\\ 
    
    \hline
    \end{tabular}
\end{center}
\end{table}{}

\subsubsection{Manipulation}
Since 2012 \citep{bagnell_integrated_2012}, BTs have been used in academia for control of robotic arm manipulation tasks, where they have proved to be useful, mainly due to the key characteristics of \emph{transparency} and \emph{modularity}. 
An example of a BT for controlling a mobile manipulator can be found in Figure~\ref{rovida17}.
\par
Within industrial robotics, BTs can be used to improve the \emph{reactivity} of systems, which is especially important for collaborative robotics.
One example of this is the commercial software platform Intera\footnote{\url{https://www.rethinkrobotics.com/intera}
} of Rethink Robotics, now owned and developed by HAHN Robotics. 
\par
Other hardware setups with collobarative robots used in publications on BTs include the ABB YuMi~\citep{colledanchise_towards_2019,csiszar_behavior_2017}, various versions of KUKA collaborative robots~\citep{colledanchise_towards_2019,rovida_extended_2017,rovida_motion_2018,paxton_costar_2017}, the Universal Robots UR5~\citep{rovida_extended_2017,guerin_framework_2015,paxton_user_2017,paxton_evaluating_2018,paxton_costar_2017,berenz_playful_2018}, the Softbank NAO~\citep{tumova_maximally_2014,marzinotto_towards_2014,colledanchise_how_2017,berenz_playful_2018}, and the Franka Emika Panda~\citep{paxton19}. There are also some examples using other robots such as the Barrett WAM~\citep{bagnell_integrated_2012} and the IIT R1 robot~\citep{colledanchise_analysis_2019}. 
\par
The majority of the BT implementations used for robotic manipulation have been constructed manually but there are also examples using \emph{Linear Temporal Logic}~\citep{tumova_maximally_2014} and~\citep{marzinotto_flexible_2017}, \emph{Backchaining} and STRIPS planners~\citep{colledanchise_towards_2019}, \emph{Planning Domain Definition Language (PDDL) planners}~\citep{rovida_extended_2017} and the \emph{A*-like planner} of~\citep{paxton19}.
Most of these papers use robotic manipulation as the main application, but the results are often applicable to robotics in general. Simulated systems are used in~\citep{colledanchise_performance_2014,rovida_extended_2017,colledanchise_towards_2019,colledanchise_analysis_2019,paxton19},
and real ones in~\citep{bagnell_integrated_2012,marzinotto_towards_2014,tumova_maximally_2014,guerin_framework_2015,paxton_costar_2017,paxton_user_2017,paxton_evaluating_2018,rovida_extended_2017,rovida_motion_2018,colledanchise_how_2017,csiszar_behavior_2017,berenz_playful_2018,paxton19}. 
\par
Looking at the timeline, the first paper using robotic arms~\citep{bagnell_integrated_2012} made a general proof of concept of task structuring with perception, planning and grasping for a Barrett WAM.
\citep{marzinotto_towards_2014} introduced a formal description of BTs and demonstrated grasping and transporting objects with NAO robots. 
\citep{colledanchise_performance_2014} described how to compute performance measures for stochastic BTs for a robotic search and grasp task in simulation.
\citep{tumova_maximally_2014} described a way to interface BTs with Linear Temporal Logic which was then demonstrated with NAO robots grasping and transporting objects. 
A series of papers,~\citep{guerin_framework_2015,paxton_user_2017,paxton_evaluating_2018,paxton_costar_2017} used the software COSTAR, with a graphical interface for creating BTs. The effectiveness CoSTAR was then demonstrated on a number of systems. \citep{guerin_framework_2015} used CoSTAR with a Universal UR5 performing a kitting task and a machine tending task, both with  manually created BTs. \citep{paxton_costar_2017} extended \citep{guerin_framework_2015} with experiments on a KUKA LBR iiwa and UR5 robot performing tasks such as wire bending, polishing, and pick and place. \citep{paxton_user_2017} and \citep{paxton_evaluating_2018} presented usability studies where users created BTs for UR5 robots in CoSTAR to perform pick and place tasks. They concluded that users \say{found BTs to be an effective way of specifying task plans}.
\citep{rovida_extended_2017} presented extended BTs (eBT), which also include conditions to enable optimization which was shown to reduce execution time on a robotic kitting task. The demonstrations were performed on a KUKA LWR4+ on a mobile platform in simulation and on a real stationary Universal Robot UR5.
\citep{rovida_motion_2018} built on~\citep{rovida_extended_2017} and described using parallel superimposed motion primitives to generate complex behaviors and used Allen's interval algebra to define constraints in spatial relations demonstrated on a robotic assembly task with a KUKA iiwa 7 R800.

\citep{csiszar_behavior_2017} used a version of BTs with support for event handling to demonstrate real assembly of a planet gear carrier with the industrial ABB YuMi robot.
\citep{colledanchise_how_2017} mainly showed theoretical results of how BTs modularize and generalize other architectures. These properties were demonstrated on NAO robots grasping and moving balls. 
 
\citep{colledanchise_towards_2019} described how to use a planning algorithm as a basis to iteratively and reactively generate BTs. The generated trees were demonstrated in two simulated scenarios. The first consisted of a KUKA Youbot doing pick and place tasks in the presence of obstacles and external agents. The second scenario was an ABB YuMi robot performing a simplistic version of cell phone assembly.
\citep{colledanchise_analysis_2019} defined and analyzed two synchronization techniques to handle concurrency problems with BTs using parallel nodes. The techniques were demonstrated on simulated ITT R1 robots in four experiments with different navigation and object pushing and manipulation tasks.
Finally, \citep{paxton19} drew inspiration from BTs and introduced a framework called Robust Logical Dynamical Systems (RLDS), shown to be equivalent to BTs in the sense that any BT can be descripted as an RLDS and vice versa. It was  shown together with a simple planner to produce robust results for a Franka Emika Panda robot grasping objects and placing them in a kitchen drawer. 

\subsubsection{Mobile Ground Robots}

\begin{figure}
  \centering
    \includegraphics[width=0.48\textwidth]{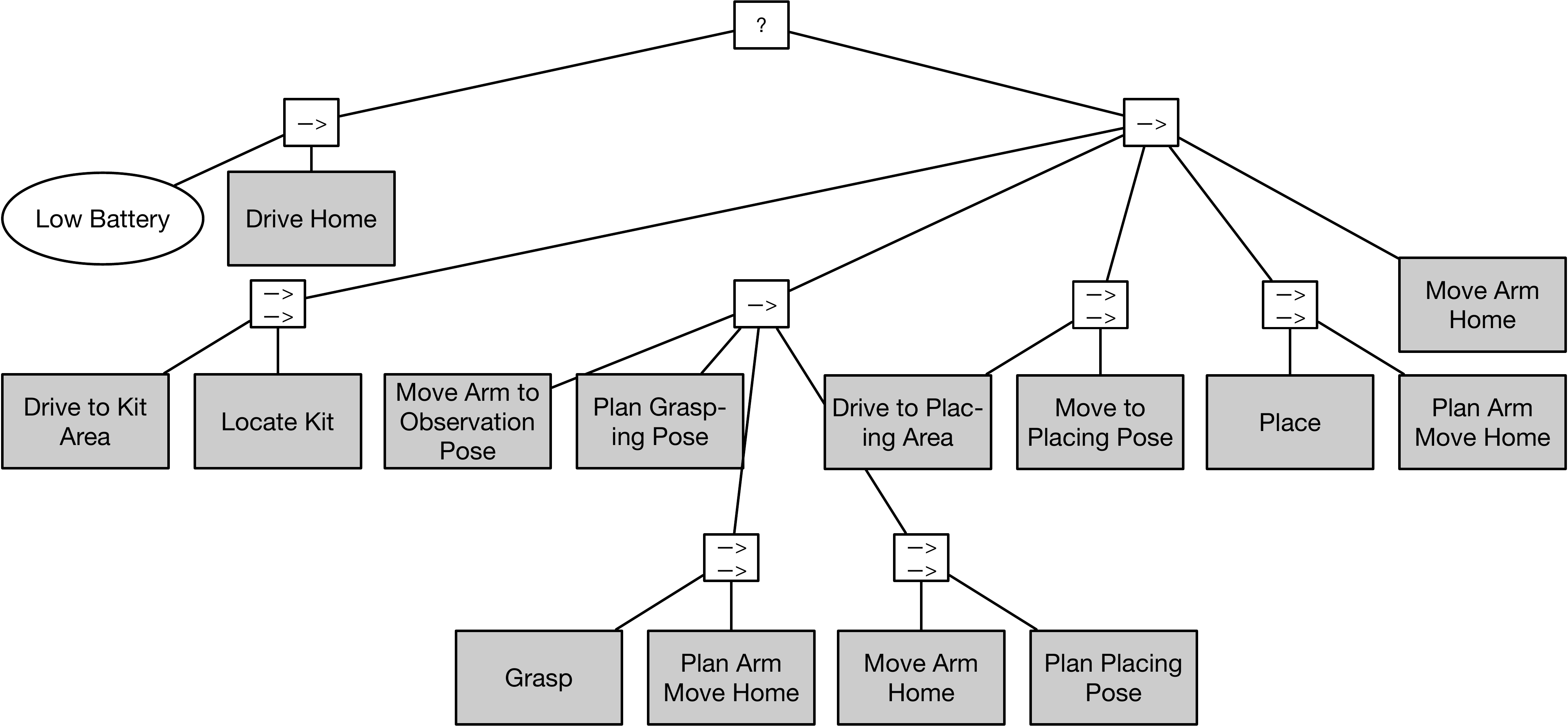}
      \caption{ Adapted from \citep{rovida_extended_2017}. A BT for a mobile manipulator to pick and place objects in a kitting task. The BT was optimized for completion time by applying an algorithm that added Parallel nodes where possible.}
      \label{rovida17}
\end{figure}

Here we describe the application of BTs for mobile ground robots. The first work in this area was~\citep{marzinotto_towards_2014,colledanchise_how_2014}, and an example of a BT for mobile manipulation can be found in Figure~\ref{rovida17}.

The \emph{reactivity} provided by BTs is useful for providing ground robots with responses to detected faults or unknown obstacles~\citep{colledanchise_advantages_2016,segura-muros_integration_2017,colledanchise_analysis_2019,colledanchise_towards_2019}, and the advantages are  similar to the ones found in manipulator control, especially for mobile manipulators, as there is significant overlap between the topics.
 
 In swarm robotics, a number of studies have been made on the combination of BTs and evolutional algorithms~\citep{parker_multiple_2008,kuckling_behavior_2018,jones_two_2018,jones_evolving_2018,neupane_emergence_2019,neupane_designing_2019,yang_hierarchical_2019}.
  The tree structure of BTs gives good support for straight-forward implementation of both mutations and cross over, while the \emph{modularity} increases the chances of such variations resulting in functional designs.

BTs are used to perform search and rescue tasks and navigation in cluttered environments in~\citep{colledanchise_advantages_2016,colledanchise_improving_2018}. Multi-robot systems are investigated in~\citep{colledanchise_advantages_2016}, where the 
\emph{Parallel} node is used to improve the inherent fault tolerance of multi-robots applications. Mission performance is formally analyzed in terms of \emph{minor} and \emph{major} faults, where the robot is not able to perform some task or any task at all, respectively. 

Fault detection is also a key characteristic of the framework proposed in~\citep{segura-muros_integration_2017} where the BT, used to control a simulated mobile manipulator, frequently checks the correspondence between the expected logic state of the world and what is sensed, allowing for replanning if discrepancies are found. Finally, in~\citep{colledanchise_analysis_2019}, a mobile manipulator (the R1 robot from IIT in Genova) is used as a proof of concept to illustrate how the execution of a parallel node can be synchronized or un-synchronized. The same robot is used in~\citep{giunchiglia_conditional_2019} in an example of a task solved by the proposed Conditional Behavior Tree.

The \emph{modularity} of BTs is crucial in applications like mobile manipulation, where the grasping capabilities of a manipulator are enhanced by a mobile platform. In particular, BTs of that kind are designed to reach a goal, subject to the condition of moving in a collision-free trajectory. This task is simulated with a KUKA Youbot in~\citep{colledanchise_towards_2019}, as a practical proof of the scalability of automatically generating BTs for solving complex problems. A robot similar to the KUKA KMR iiwa is used in~\citep{rovida_extended_2017} to show that a BT, combined with an algorithm optimizing a skill sequence generated by a PDDL planner, can save  20\% of the execution time, compared to a standard sequential execution. Similarly, in~\citep{zhou_autonomous_2019}, a dual-arm mobile manipulator takes user's directions as audio input and uses automatic generated BTs to perform a picking task. In~\citep{french_learning_2019} a mobile manipulator performs a household cleaning task where navigation is required, using a BT learned from experience, as an alternative to the Decision Trees used in previous works. The paper uses BTs as a policy representation for the task because of its \emph{modularity}, \emph{transparency} and responsiveness. In~\citep{jiang_laair_2018,kim_architecture_2018} BTs are used to model a person following task, executed by a Toyota HSR (Human Support Robot). Here, the responsiveness of the BT is fundamental for coordinating sensor data with robot skills, to react to the changes of a dynamic environment. 

BTs are combined with Semantic Trajectories (a trajectory enriched with context information) in~\citep{siqueira_context-aware_2015} and~\citep{siqueira_semantic_2016}. This framework is applied to the navigation of a X80SV mobile robot performing a patrol task in simulation. BTs are used to implement the decision making algorithm, which uses a scenario defined by a set of context information provided by the semantic trajectories. In these papers, the authors show that the trajectory enriched framework improves the task execution of the autonomous agent. In~\citep{banerjee_autonomous_2018}, a simulated iRobot Create 2 performs a package delivery task. Here, Reinforcement Learning is used to learn a policy which is then automatically converted into a BT, which is stated to \say{offer an elegant solution [\ldots to \ldots] the need for uniform action duration in both planning and plan recognition}. Another example is portrayed in~\citep{macenski_marathon_2020}, where the authors present \emph{Navigation2}, a ROS2 realization of the ROS Navigation Stack, which features configurable BTs to handle planning, control, and recovery tasks. The framework is tested in dynamic environments using TIAGo and RB-1 robots.

BTs have also been applied in decision making for soccer robots in~\citep{abiyev_control_2013,abiyev_robot_2016}, to control wheeled holonomic robots in the RoboCup competition. In particular, \citep{abiyev_control_2013} combines the decision output of the BT (e.g. whether to pass, shoot or move) with a RRT path planner to achieve robot movement. In \citep{abiyev_robot_2016}, the same authors combine BTs with Fuzzy logic and use Fuzzy membership functions as the leaves of the BT. A Fuzzy obstacle avoidance algorithm is then used for path planning. Experiments show that the Fuzzy logic obstacle avoidance algorithm has a better performance in terms of time and length of the computed path, when compared to other planners based on RRT or A$^*$. They emphasize how the \emph{modularity} of BTs \say{allows to easily extend decision making for complex tasks} and simplify maintainability.

In Swarm Robotics a multi-robot system composed of a large number of homogeneous robots re-creates behaviors that are often inspired by biology (e.g. ants)~\citep{parker_multiple_2008}. Some examples in the literature use e-Puck or Kilobot swarms controlled by BTs created using evolutionary algorithms.
In~\citep{kuckling_behavior_2018} a robot swarm is composed of twenty 2-wheeled e-puck robots, whose behavior is governed by BTs with a restricted topology that are automatically generated by the Maple planner. The hardware  of the robots limits the pool of leaves to six actions and six conditions. The tasks executed by the swarm are \emph{foraging} and \emph{aggregation}. A similar swarm of sixteen e-pucks is used in~\citep{jones_two_2018} and controlled by evolved BTs to push a frisbee in an arena. The same authors use evolved BTs to control a swarm of Kilobots to carry out a foraging task in~\citep{jones_evolving_2018}. Here, the BT approach is preferred because of its \emph{transparency} to a human observer. Finally, in~\citep{neupane_emergence_2019, neupane_designing_2019} BNF grammar implementing the Grammatical Evolution genotype-to-phenotype transformation incorporates rules that produce BTs to represent swarm behaviors. In this work, evolved behaviors are shown to perform better than a hand-coded one for a swarm performing single source foraging, nest maintenance and cooperative transportation. Robotic swarm control using BTs was also done in~\citep{yang_hierarchical_2019}, as an execution module in a hierarchical framework.

Concerning walking robots, in~\citep{marzinotto_towards_2014,tumova_maximally_2014}, BTs are used to make the Softbank NAO robot walk towards and grasp a target object. The robot application is used as a proof of concept for the BT framework. In~\citep{hannaford_simulation_2016}, BTs are used to adapt the walking capabilities of a simulated humanoid robot navigating through three different types of terrain. In this latter case, the \emph{Fallback} node is exploited and it learns how to change the step depth of the robot depending on the terrain conditions. In~\citep{coronado_robots_2019}, a hand-coded BT controls a Pepper robot for CRI (Child-Robot Interaction) \say{in the wild} experiments. A graphic block-based interface allows non-programmers to generate complex robot behaviors powered by BTs.

Finally, a research field in which BTs have not yet been applied to a large extent is Autonomous Driving. In most cases, as described in~\citep{paden_survey_2016}, the decision layer of an autonomous vehicle is a Finite State Machine (FSM) or Partially Observed Markov Decision Process (POMDP). Thus, given that BTs generalize these structures, an autonomous vehicle can also be controlled by a BT, and the advantages in terms of \emph{modularity}, \emph{reactivity} and \emph{transparency} could be useful in unpredictable urban environments as well as highway scenarios. This is loosely suggested in~\citep{colledanchise_how_2014}, where a simple example includes a car performing lane following, overtaking and parking. These behaviors have also been simulated in~\citep{olsson_behavior_2016} using both  FSMs and BTs. It is  shown how the latter scale better when compared to the former, when more behaviors are added, allowing increased reusability and structure simplicity.


\subsubsection{Aerial and Underwater Robots}

\begin{figure}
  \centering
    \includegraphics[width=0.35\textwidth]{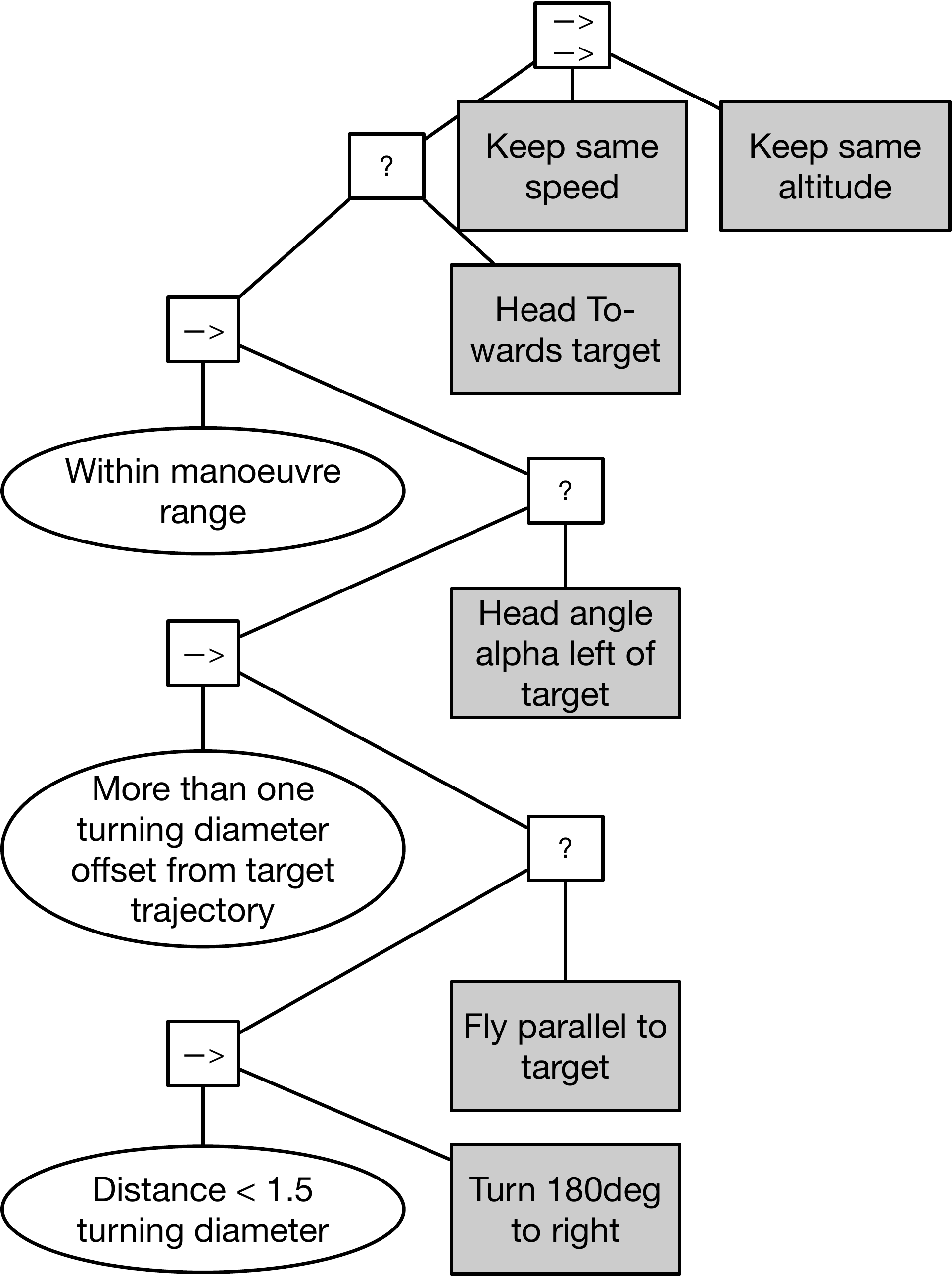}
      \caption{ Adapted from \citep{ramirez_integrated_2018}. A BT for UAVs to move towards and fly in behind another UAV.}
      \label{ramirez18}
\end{figure}

BTs have been used for two categories of Unmanned Aerial Vehicles (UAVs), small multi-copters moving in close proximity of obstacles~\citep{scheper_behavior_2015,lan_modular_2018} and large fixed wing aircraft moving at higher altitudes in longer missions~\citep{klockner_behavior_2013,ramirez_integrated_2018}.

For small UAVs, BTs were used in the winning design of the 2017 International Micro Aerial Vehicle Competition (IMAV 17)~\citep{lan_modular_2018}. There, the UAV was tasked with moving through a window, search for an object, pass in front of a fan, find a location and drop an object there and finally land on a moving platform. The advantages of BTs in terms of \emph{modularity} and hierarchical structure was mentioned as key elements of the approach used. A related task was addressed in~\citep{scheper_behavior_2015} using evolutionary methods. Finally, manually designed BTs were used as a key part of a building inspection drone design in~\citep{li_uav_2019}, and an LTL planning approach was used for a UAV search mission in~\citep{lan_autonomous_2019}. In~\citep{goudarzi_mission_2019} an increase in automation was achieved by implementing BT's for UAV systems that operate in congested areas. 

For larger UAVs, BTs were used to do mission management for extremely high altitude operations in a series of papers~\citep{klockner_behavior_2013,klockner_modelica_2014,klockner_behavior_2015,klockner_interfacing_2018}, that include real, practical operations.
BTs have also been proposed for the control of a UAV chasing other aircraft~\citep{ramirez_integrated_2018} as illustrated in  Figure~\ref{ramirez18}.

Finally, BTs have been used to address the strict safety requirements that are associated with the operation of large UAVs.
Issues regarding synchronization and memory were investigated to enable redundancy for BTs in~\citep{safronov_asynchronous_2019},
and BTs for fault detection and mitigation were proposed in~\citep{castano_safe_2019}.

The operation of Autonomous Underwater Vehicles (AUVs) also requires very reliable systems, not to avoid damage to bystanders, but rather to reduce the risk of loosing the vehicle itself. Thus, a BT approach for AUV mission management, including fault handling, was proposed in~\citep{sprague_improving_2018}.

\subsection{Other}

Behavior trees have also been used for a large variety of applications apart from those listed in previous sections.

\begin{table}[]
    \caption{Other Applications}
    \label{tab:others}
\begin{center}
    \begin{tabular}{|p{2.5cm}|p{4.5cm}|}
    \hline
    Area & Papers  \\ \hline
    Workflow and human activity modelling & \citep{deneke_conceptual_2017},\citep{hu_semi-autonomous_2015}, \citep{hannaford_behavior_2018}, \citep{bergeron_rfid_2017}, \citep{hu_semi-autonomous_2018}, \citep{bouchard_modeling_2018}, \citep{shu_behavior_2019}\\
    \hline
    Camera Control & \citep{prima_secondary_2013}, \citep{markowitz_intelligent_2011} \\
    \hline
    Smart homes and Power Grids & \citep{francillette_towards_2016}, \citep{brusco_compact_2018}, \citep{burgio_compact_2018}, \citep{haijun_simulation_2019} \\
    \hline
    Engineering Design Tasks &  \citep{zhang_ikbt_2019}, \citep{parfenyuk_development_2015} \\
    \hline
    \end{tabular}
\end{center}
\end{table}{}

Since a BT represents an action policy, they can be used as a general language for describing workflows and procedures to completely different types of tasks. One such class of tasks is found in medicine. Using BTs for online procedure guidance for emergency medical services has been proposed in~\citep{shu_behavior_2019}, in a manner similar to dialogue systems for games. Surgical procedures have also been formulated in BT notation, in an agent-agnostic way that allows the same description to be used for procedures to be carried out either manually by the surgeons themselves or by a robot~\citep{hannaford_behavior_2018}. Other examples have specifically targeted joint human-robot task execution. In one approach, human interaction was used to get input for a Fallback node, to take advantage of an experienced operator's (brain surgeon) expertise in choosing the correct action for a brain tumor ablation task. This has been demonstrated in both simulation~\citep{hu_semi-autonomous_2015} and in real experiments on mice~\citep{hu_semi-autonomous_2018}.

Apart from using BTs to prescribe human tasks, they have also been used to describe observed actions and task executions. A conceptual model for how to represent general human workflows using BTs was presented in~\citep{deneke_conceptual_2017}, along with a prototype demonstration of how to encode cooking procedures. In~\citep{bergeron_rfid_2017}, activities of daily living have been observed as interaction constraints between tracked objects, which are then automatically converted into BT descriptions. Similar descriptions are used to encode activities for daily living in~\citep{bouchard_modeling_2018}, where the demonstrated implementation is generating behaviors for virtual agents in a simulated home environment.
Camera placement for animation production in virtual environments can also be controlled with BTs~\citep{markowitz_intelligent_2011,prima_secondary_2013}.

Furthermore, the BT framework has been used to formalize procedures for problem solving. One example is to find closed form inverse kinematics for a general serial manipulator~\citep{zhang_ikbt_2019}. The procedure carried out by humans is encoded in a BT and the authors argue that the same approach can be applied to other complex but well-structured cognitive tasks. Another example is the design of power grid systems, where the design procedure is automated in a similar way using BTs~\citep{parfenyuk_development_2015}. 

Finally, apart from encoding task execution for autonomous agents and humans, BTs have been also been used to run simpler automated systems, such as smart homes. The \emph{reactivity} of the BTs can be exploited to control a domestic power grid, by balancing loads and engaging local back-up power production when needed~\citep{brusco_compact_2018,burgio_compact_2018,haijun_simulation_2019}, or control the activation of lighting or smart appliances based on human actions~\citep{francillette_towards_2016}.

\section{Methodology}
\label{sec:method}
BTs were originally conceived as a tool to manually create modular task switching structures. However, as \emph{modularity} is beneficial for both manual and automatic synthesis, 
BTs have been successfully used as a policy representation for both learning and planning approaches, with the hope of improving both performance and design time.

\subsection{Learning}
\label{sec:learning}

The learning approaches to designing BTs can be divided into Reinforcement Learning, Evolution-inspired learning, Case Based Reasoning and Learning from demonstration, as seen in Table  \ref{tab:learning}. Some of these methods have been applied to automatically generate BT's while other methods have been used to improve existing parameters within a predefined BT. Below we discuss different learning approaches in more detail.

\begin{table}[]
 \caption{Learning approaches to synthesising a BT.}
    \label{tab:learning}
\begin{center}
    \begin{tabular}{|p{3.2cm}|p{4.5cm}|}
    \hline
    Learning Method & Papers  \\ \hline
    Reinforcement &   \citep{dey_ql-bt_2013},
    \citep{fu_reinforcement_2016}, \citep{kartasev_integrating_2019},  \citep{banerjee_autonomous_2018}, \citep{zhang_modeling_2017},
    \citep{zhang_combining_2017}, \citep{holzl_reasoning_2015}, \citep{holzl_continuous_2015}, \citep{pereira_framework_2015},  \citep{fu_reinforcement_2016}, \citep{zhang_towards_2016},
    \citep{zhu_behavior_2019}\\
    Evolutionary & \citep{hoff_evolving_2016}, \citep{hutchison_evolving_2010}, \citep{perez_evolving_2011}, \citep{estgren_behaviour_2017}, \citep{paduraru_automatic_2019}, \citep{zhang_behavior_2018}, \citep{nicolau_evolutionary_2017}, \citep{yao_adaptive_2015}, \citep{eilert_learning_2019}, \citep{berthling-hansen_automating_2018}, \citep{neupane_learning_2019}, \citep{schwab_capturing_2015}, \citep{jones_two_2018}, \citep{neupane_emergence_2019}, \citep{neupane_designing_2019}, \citep{jones_two_2018}, \citep{jones_evolving_2018}, \citep{colledanchise_learning_2019},
    \citep{schwab_capturing_2015}, \citep{neupane_emergence_2019}, \citep{yao_tactics_2017},
    \citep{hallawa_evolving_2020}\\
    Case Based Reasoning & \citep{florez-puga_dynamic_2008}, \citep{palma_extending_2011}, \citep{florez-puga_empowering_2011}, \citep{zhang_towards_2016} \\
    From Demonstration & \citep{french_learning_2019}, \citep{sagredo-olivenza_combining_2017},
    \citep{buche_orion_2020}\\
    \hline
    \end{tabular}
\end{center}
   
\end{table}{}

\subsubsection{Reinforcement Learning (RL)}

In \citep{dey_ql-bt_2013}, an existing BT  is  re-ordered using Q-values that are computed using the lowest level Sequences as actions in the RL algorithm.
In  \citep{fu_reinforcement_2016}, the RL takes place in the Fallback nodes, executing the child with the highest Q-value.
Similar ideas were built upon in  \citep{zhang_combining_2017} combining the hierarchical RL approach MAXQ with BTs. Starting from a manual design, some Fallback nodes are marked to be improved by learning. Having found the Q-values of each child, new condition nodes were created to allow execution of each child when that child is the best option. This line of ideas was also explored in \citep{zhu_behavior_2019}.
The idea of combining RL and BTs was also discussed at a more general level in \citep{holzl_reasoning_2015,holzl_continuous_2015}.
Individual actions or conditions can also be endowed with RL capabilities, a concept that was explored in~\citep{kartasev_integrating_2019,pereira_framework_2015}.
The later specified learning nodes in the BT where an RL problem in terms of states, actions and rewards was defined.
Finally, an approach involving the translation of a learned RL policy into a BT was suggested in~\citep{banerjee_autonomous_2018}.

\subsubsection{Evolution-inspired Learning}

\begin{figure}
  \centering
    \includegraphics[width=0.38\textwidth]{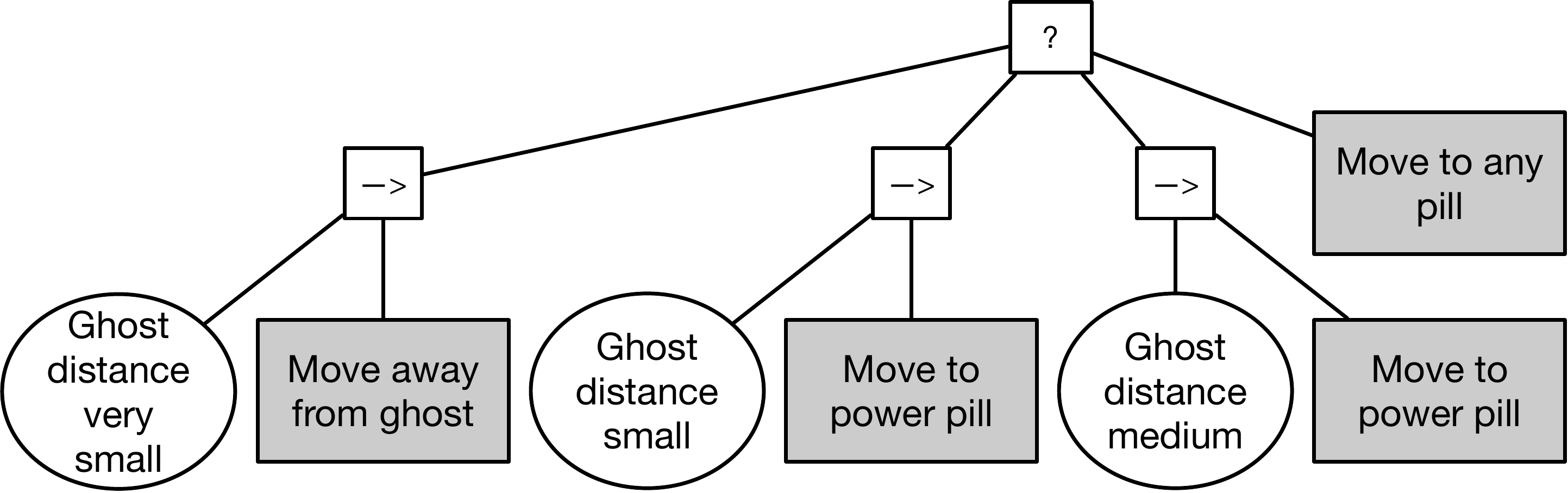}
      \caption{ A BT for Pacman, that is learned using evolutionary methods, adapted from \citep{zhang_learning_2018}. }
      \label{zhang18}
\end{figure}

Evolutionary algorithms build upon the ideas of having a population of solution candidates divided into generations. For each generation, some kind of fitness evaluation is performed to select a subset of the generation, then a new generation is created from this subset by applying operations such as mutations and crossover.
As seen below, BTs are very well suited to such algorithms.
It has been shown that \emph{locality}, in terms of small changes in design giving small changes in performance, is important for the performance of evolutionary algorithms \citep{rothlauf2006locality}.
Thus, it seems that the \emph{modularity} of BTs is providing the locality needed by these algorithms.

Genetic programming is one form of evolutionary algorithms that has been applied to BTs, see Figure~\ref{zhang18}, and a number of studies have shown that solutions that outperform manual designs can be found~\citep{hoff_evolving_2016,hutchison_evolving_2010,perez_evolving_2011,neupane_emergence_2019,colledanchise_learning_2019}. These methods have been used to automatically evolve new behaviors using high-level definitions given a new environment~\citep{estgren_behaviour_2017,paduraru_automatic_2019} and additional improvements have been made to increase scalability for evolving algorithms~\citep{zhang_behavior_2018}. A combination of grammatical evolution with BTs and path-finding algorithms such as A* was explored in~\citep{nicolau_evolutionary_2017}. 
Air Combat simulators are  highly dynamical environments where human behavior is often represented by rule-based scripts. Grammatical evolution has been used to create adaptive human behavior models~\citep{yao_adaptive_2015} and studies have been done to improve pilot decision support during reconnaissance missions~\citep{eilert_learning_2019}. Other military-related training simulators included ground unit mobility and training on how to act in dangerous situations. To reduce the need for military experts to explain to programmers how virtual soldiers  should behave in a particular situation, a genetic algorithm has been used to automatically generate such agents~\citep{berthling-hansen_automating_2018}. 
In multi-agent cases, it is often difficult to specify policies for every single agent and thus the same hand-written BTs have been used for many units. Using an evolutionary approach, individual BTs were developed for each agent, resulting in a more natural and efficient behavior of the whole team~\citep{neupane_designing_2019,neupane_learning_2019}, but the downside of such automated techniques is reduced designer control~\citep{schwab_capturing_2015}. Another example of a multi-agent system is a robotic swarm. In this field, BTs have been generated using both grammatical evolution~\citep{neupane_emergence_2019} and genetic programming. In~\citep{jones_two_2018,jones_evolving_2018} the evolutionary system were based on a  \emph{distributed island model}.
Some of the solutions generated by Evolutionary Algorithms can be quite large and complex. Both \citep{colledanchise_learning_2019} and \citep{hallawa_evolving_2020} describe methods for reducing this problem.

\subsubsection{Case Based Reasoning}
Case Based Reasoning is an automated decision making process where solutions to new problems are provided through experiences from problems that were previously encountered. Following this process, rules can be constructed that may solve future problems by retrieving and reusing stored behaviors~\citep{florez-puga_dynamic_2008,palma_extending_2011}. One drawback of this method has been the difficulty of refining existing strategies that have been previously learned from experience~\citep{palma_combining_2011}, but it has been shown that some additional adaptability can be provided using RL~\citep{zhang_towards_2016}.

\subsubsection{Learning from Demonstration}
Usually one needs an expert or a person with good knowledge in the field of robotics to program a robot for executing a particular task. Much less knowledge is required if the robot can learn from a situation where the human demonstrates a particular task that the robot needs to execute~\citep{french_learning_2019}. Similarly, creating an NPC for a game can be a time consuming process, where game designers have to interact with a programmer for creating the desired characters. Learning by demonstration has shown improvements in generating more human-like behavior, although in situations where human behavior was less predictable, the resulting NPC behaviors might not meet expectations~\citep{buche_orion_2020}. Machine learning functions, such as neural networks, have been used in combination with BTs to let game developers combine hand-coded BTs with programming by demonstration~\citep{sagredo-olivenza_combining_2017}. This technique can increase performance in uncertain environments thanks to the noise tolerance of neural networks.

\subsection{Planning and Analytic design}
\label{sec:planning}

\begin{figure}
\centering
\includegraphics[width=0.5\textwidth]{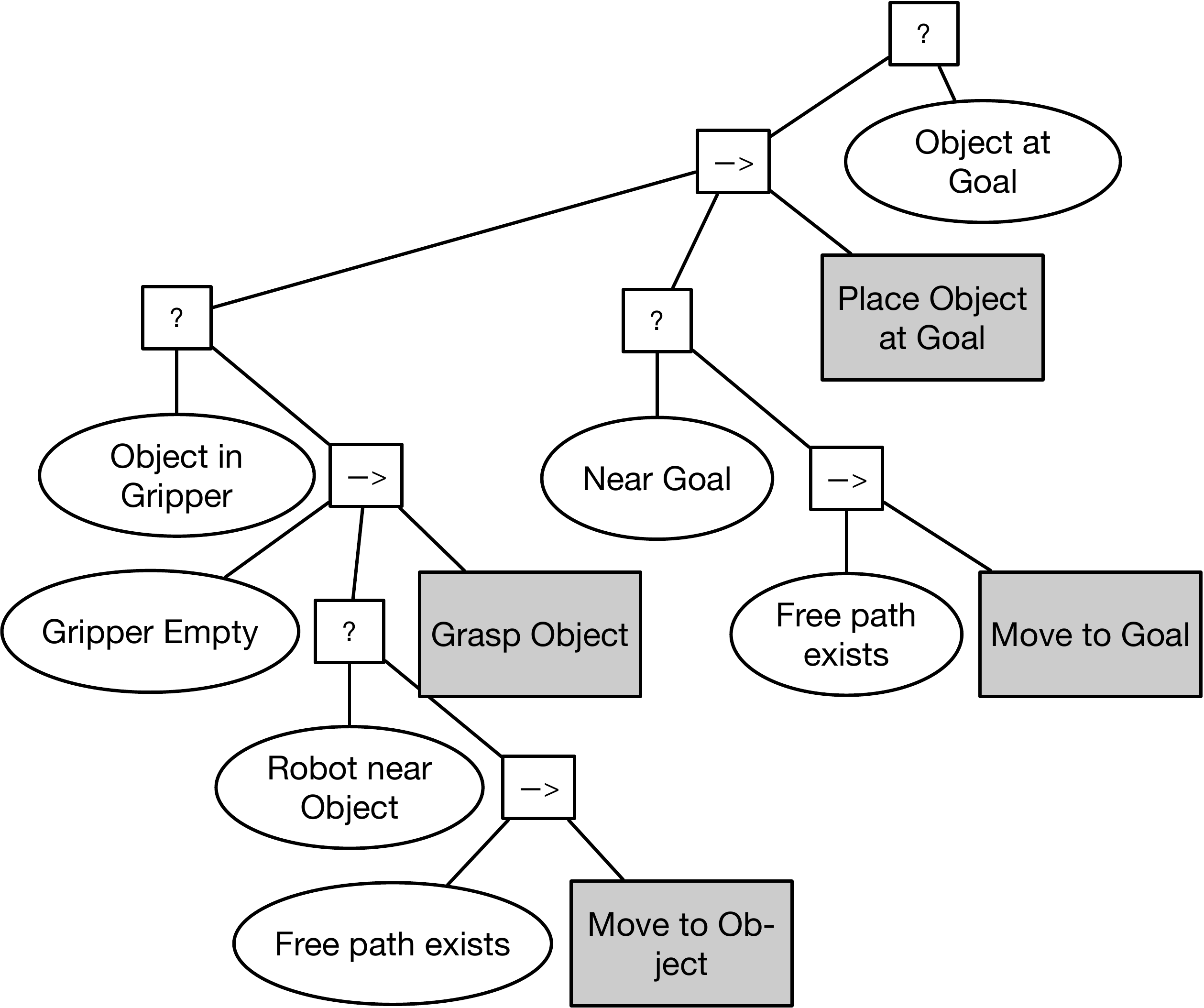}
\caption{A BT for mobile manipulation, generated using a planning approach, and data from a table of actions and preconditions, as illustrated in Table \ref{tab:BCh}. Adapted from \citep{colledanchise_towards_2019}.}
\label{colledanchise16}
\end{figure}

In this section we review the papers that show how a BT, such as the one in Figure~\ref{colledanchise16}, can be synthesized using a planner. Depending on the planner chosen, these papers can be categorized as in Table~\ref{tab:planning}.

\begin{table}[]
  \caption{Different approaches for synthesising a BT using a planning tool.}
    \label{tab:planning}
\begin{center}
    \begin{tabular}{|l|p{5.1cm}|}
    \hline
    Planner & Papers  \\ \hline
    STRIPS type & \citep{colledanchise_towards_2019} \\
    HTN & \citep{neufeld_hybrid_2018}, \citep{rovida_extended_2017}, \citep{segura-muros_integration_2017}, \citep{holzl_reasoning_2015}\\
    Auto-MoDe & \citep{kuckling_behavior_2018}\\
    LTL & \citep{tumova_maximally_2014}, \citep{marzinotto_flexible_2017}, \citep{colledanchise_synthesis_2017},
    \citep{lan_autonomous_2019}\\
    GOAP & \citep{schwab_capturing_2015} \\
    Other & \citep{tadewos_automatic_2019}, \citep{tadewos_--fly_2019}, \citep{zhou_autonomous_2019} \citep{paxton19}\\
    \hline
    \end{tabular}
\end{center}
    
\end{table}{}

Planning algorithms are generally used to create action sequences to achieve long term goals. However, they have the important limitation that if the agent fails to execute an action  or a subtask, it has to re-plan. 
Such failures are usually due to uncertainties. Perhaps the outcome of an action was different than expected, or the state of the world, in terms of positions and geometry of all sorts of objects and agents, was different than expected.
Thus, the more uncertain, changing and dynamic the world is, the more often plans will fail. Also, re-planning takes time and computational resources, especially if the task is complex and involves many objects and actions. 

BTs are inherently reactive, and by constantly checking conditions they switch tasks based on the perceived state of the world. However, the size of the BT must be very large to perform tasks that involve many objects and subtasks, and the larger the BT, the more difficult it is to design manually. The idea of using planning algorithms to construct BTs adresses this problem, trying to combine the \emph{reactivity} of BTs with the goal directed nature of plans. 

The most common approach to synthesise a BT from a planner consists of two steps: first, the planner computes a plan to solve a given task, then a planner-specific algorithm converts the plan into a BT which is finally used to control the agent (a robot in many cases). 

Several authors have proposed to generate BTs from a Hierarchical Task Network (HTN), as a natural extension of STRIPS-style planners, probably because this type of planning results in hierarchical task decompositions which have a natural mapping to the hierachical structure of BTs. 

In~\citep{colledanchise_towards_2019}, the authors combine BTs, relying on their \emph{reactivity} and \emph{modularity}, with a STRIPS-style planning algorithm, leveraging the idea of backchaining: they start from the goal condition and proceed backwards, iteratively finding the actions that meet that goal condition. Input to the problem is a set of actions with the corresponding preconditions and postconditions (Table \ref{tab:BCh}).

Furthermore, it is reactive to changes in the environment brought by external agents and it can be expanded during runtime.

Based on the same idea,~\citep{tadewos_automatic_2019} and~\citep{tadewos_--fly_2019} present their own algorithm for the automatic synthesis of BTs. This algorithm is combined with dynamic differential logic (DL), a verification tool used in~\citep{tadewos_automatic_2019} to ensure the safety of the generated BT. Then, DL is combined with a market-based auction algorithm and a coordinator to assign specific task to certain robots, in the multi-system scenario presented in~\citep{tadewos_--fly_2019}.
In \citep{paxton19} a heuristic planner is used to create BTs in the form of Robust Logical-Dynamical Systems (RLDS).

\begin{table}
\centering
\caption{Example of an input to the problem addressed in \citep{colledanchise_towards_2019}.}
\begin{tabular}{c|c|c} 
 Actions & Preconditions & Postconditions \\ 
\hline
$A_1$ & $C_{11}^{Pre}, C_{12}^{Pre}$ & $C^{Post}$ \\ 
$A_2$ & $C_{21}^{Pre}, C_{22}^{Pre}$ & $C^{Post}$ \\
\hline
\end{tabular}
\label{tab:BCh}
\end{table}

In~\citep{neufeld_hybrid_2018}, a Game AI is built using HTN planners alongside BTs. The main contribution of this work is the proposition of an hybrid approach where HTN is used to compute high-level strategic plans while the BTs are managing low-level tasks. Despite the presence of the hybrid control strategy combining the planner with BTs, the BTs are still hand coded.
An automatic transition from HTN to BTs is proposed in~\citep{rovida_extended_2017}. Here, the pre- and post-conditions of HTNs are added to BTs, in a new framework called eBT (extended BT). In particular, the HTN is initialized with a planner based on a PDDL (Planning Domain Definition Language) representation, translating a semantic model of the world, shared among all the nodes in the eBT. The overall procedure is accomplished in two phases: first the PDDL planner generates a sequence of skills (actions), then an eBT is generated and optimized for execution. 

HTNs are combined with BTs and implemented in ROS in~\citep{segura-muros_integration_2017}. Here, the HTN planner computes a plan from input tasks, which is automatically converted into a BT that executes it. Also, the user is free to define task-dependent subtrees which are automatically dealt with and combined with the pre-existing BTs. In this framework, the planning and execution layer communicate through a blackboard which is also responsible for tracking and checking the execution. If the actual current logic state does not correspond to the expected logic state,  replanning is needed. 
The extended BT (XBT) proposed in~\citep{holzl_reasoning_2015} also uses HTN planning. It is proposed to create virtual copies of the states of the tree and evaluate it, in order to prune behaviors that either fail or do not contribute to reaching the goal.

Inspired by the \emph{precondition/postcondition} mechanism in PDDL,~\citep{zhou_autonomous_2019} combine the BT representation model with the \emph{Hierarchical task and motion Planning in the Now} (HPN) planning algorithm, to design an algorithm that synthesizes BTs automatically. The planner outputs a hierarchical sequence of post-conditions $\rightarrow$ actions $\rightarrow$ pre-conditions, which is then translated into a BT. At runtime, the algorithm also keeps track of the failing behaviors, thus allowing for replanning.

In \citep{kuckling_behavior_2018} Maple, as an instance of AutoMoDe (automatic modular design) is used to  automatically generate a restricted topology of BTs, limited to a Sequence* node\footnote{\emph{Sequence} with memory.} as a root, which has Fallback nodes as children, which in turn have one Condition and one Action leaf each. The number of subtrees rooted with the Fallback node is limited to four. Moreover, due to the robot setup, the BT leaves are chosen from a set of six actions and six conditions. Iterated F-race is used as an optimization algorithm to search for the best BT among the generated ones. The Maple method is compared with Chocolate (which, similarly, generates probabilistic FSMs) and EvoStick (which is an implementation of an evolutionary robotics approach that uses neural networks), in the tasks of \emph{foraging} (retrieve an object from a source area and place it in a nest area), and \emph{aggregation} (the swarm is asked to aggregate in an area), both in simulation and reality. In the first task {Maple} and {Chocolate} perform better than {EvoStick} both in simulation and reality. In the second task instead, {EvoStick} performs better in simulation but worse in reality when compared to {Maple} and {Chocolate}, which in turn have similar results. 
 
An alternative approach to reactive planners is proposed in~\citep{tumova_maximally_2014}. Here, the synthesis of a robot controller is a two-step process. In the first step, the action capabilities of the robot are modeled as a transition system and the goal as a State-Event LTL formula (Linear Temporal Logic). Then, an I/O automaton synthesizes the control strategy from the LTL formula, which is finally automatically implemented in a BT. In a similar scenario,~\citep{colledanchise_synthesis_2017} shows how to synthesize a BT that guarantees to satisfy a task defined as a LTL formula. The rigorous and complex formulations of the LTL formula and the I/O automaton make this approach scale poorly in more complicated tasks or less predictable environments. Moreover the BT is generated offline. LTL generated plans are also used in~\citep{lan_autonomous_2019}, where BTs refine high-level task plans expanding them into primitive actions.

Finally, \citep{schwab_capturing_2015} proposes a method that automatically generates a BT from GOAP (Goal-Oriented Action Planning). The method requires a symbolic definition of a problem, which can be solved by a GOAP algorithm, then computes the solution in three phases. First, a Monte-Carlo simulation outputs a set of solutions produced by the algorithm for different initial states. Then, the most representative sequences in this solution set are identified by building a N-Gram model from the Monte-Carlo simulation. Finally, the N-Gram model is merged through a genetic algorithm into  a BT that mimics the execution of the GOAP algorithm.


\subsection{Manual Design and other approaches}
\label{sec:manual}
The majority of the BTs used for the applications described in Section \ref{sec:appl} are designed manually. The reason for this is that BTs were created to support manual design, and the methods for synthesising BTs automatically using learning and planning are not yet mature enough to compete with the ease of manual design, especially in cases where the required BT is fairly small.
Therefore, alongside the development on automatic synthesis, there has also been work on how to support the manual design process, as seen in Table~\ref{tab:man_design}.

\begin{table}[]
    \caption{Different aspects of manual design.}
    \label{tab:man_design}
\begin{center}
    \begin{tabular}{|p{3.2cm}|p{4.5cm}|}
    \hline
    Type of Design & Papers  \\ \hline
    Design Principles & 
    \citep{colledanchise_how_2017},
    \citep{colledanchise_behavior_2018},
    \citep{colledanchise_towards_2019},
    \citep{nilsson1993teleo}\\
    Design Tools & 
    \citep{becroft_aipaint_2011},
    \citep{shoulson_parameterizing_2011},
    \citep{olsson_behavior_2016},
    \citep{guerin_framework_2015},
    \citep{paxton_costar_2017},
    \citep{paxton_user_2017},
    \citep{paxton_evaluating_2018},
    \citep{weber_reactive_2010},
    \citep{weber_building_2011},
    \citep{mateas_behavior_2002},
    \citep{berenz_playful_2018},
    \citep{coronado_robots_2019}\\
    Hybrid Approaches &
    \citep{pereira_framework_2015},
    \citep{neufeld_hybrid_2018},
    \citep{sprague_adding_2018},
    \citep{klockner_interfacing_2018},
    \citep{rovida_motion_2018} \\
    \hline
    \end{tabular}
\end{center}
\end{table}{}

Many solutions suggest hybrid approaches. In~\citep{hannaford_simulation_2016} the BTs are hand-coded but the Fallback node is endowed with the capability of learning from experience. In~\citep{pereira_framework_2015}, specific action nodes have learning capabilities (Q-learning in particular). In \citep{neufeld_hybrid_2018} hand coded BTs are used alongside the HTN planner. In \citep{sprague_adding_2018}, a BT is manually designed to include a Neural Network Controller. In \citep{klockner_interfacing_2018} BTs are manually coded to translate a formalism from the ALC(D) description logic, and in \citep{rovida_motion_2018}, BTs are manually combined with Motion Generators.

Some authors have created and used supporting tools to sketch and design the BTs. In~\citep{becroft_aipaint_2011} the sketch-based authoring tool \emph{AIPaint} is presented. The same year,~\citep{shoulson_parameterizing_2011} introduced the graphical interface \emph{Topiary}. \emph{RIZE} is another graphic interface, based on building blocks, which is used in~\citep{coronado_robots_2019} to design BTs. \citep{olsson_behavior_2016} constructed their own plugin BT editor for \emph{Unity3D}, and today there are also many others available. Lastly, the \emph{CoSTAR} user interface has been developed over a number of publications~\citep{guerin_framework_2015,paxton_costar_2017,paxton_user_2017,paxton_evaluating_2018}.

In~\citep{weber_reactive_2010} and~\citep{weber_building_2011} an Active BT method is used, which is implemented in ABL (A Behavior Language), a reactive planning language used for Game AI. One of the benefits of this language is claimed to be the scheduling of the actions, which  is handled by a planner. Note that one still has to hand-code the desired behaviors for the AI agent according to the ABL semantics, even though the transition to the ABT is automatic. ABL is used also in~\citep{mateas_behavior_2002,simpkins_towards_2008}.

When doing manual design, inspiration can be drawn from the fact that BTs have been shown to generalize a number of other control architectures,~\citep{colledanchise_how_2017}. Such design principles are discussed in detail in~\citep{colledanchise_behavior_2018}. For some problems, the choice of action is naturally done in a Decision Tree fashion, checking conditions on different levels to finally end up in a suitable leaf of the tree. Note that Decision Trees have no return status, so it is a special case of general BT design. For other problems, a so-called implicit sequence design, inspired by the Teleo-Reactive approach~\citep{nilsson1993teleo} might be better, or a backward chained approach as suggested in~\citep{colledanchise_towards_2019}.

Another use of manually designed BTs is proposed in~\citep{zhang_ikbt_2019}, where a hand-coded BT is used to solve the Inverse Kinematics of a robot, up to 6 DoF, taking as input the Denavit-Hartenberg parameters. The outputs of this application are Python and C++ code as well as a report of the results in LaTex.

\section{Implementation}
\label{sec:impl}

In this section we give an overview of some of the most common libraries available to implement BTs. We also list the implementation language, the presence of a GUI, whether they can communicate with ROS, and whether they are open source. Moreover, we report some data from the {Github} repository to give the reader a sense of the usage of the library. Please note that this list is not complete, as there is a very large set of alternatives out there.

\begin{table*}
\centering
\caption{Some of the BT libraries available on November 2019.}
\label{tab:lib}
    \begin{tabular}{|l|l|m{6mm}|m{6mm}|m{8mm}|l|l|l|}
    \hline
    Name & Language & GUI & ROS & Open source & Forks & Stars & Last Commit\\
    \hline
        py\_trees\footnote{\url{https://github.com/splintered-reality/py_trees}} & Python &  & \checkmark & \checkmark & 36 & 87 & \emph{up to date}\\
        Owyl\footnote{\url{https://github.com/eykd/owyl}} & Python &  &  & \checkmark & 13 & 60 & Nov. 29, 2014\\
        Playful\footnote{\url{https://playful.is.tuebingen.mpg.de/}} & Python & & \checkmark & \checkmark & 0 & 3 & Nov. 1, 2019\\
        behave\footnote{\url{https://github.com/fuchen/behave}} & Python & &  & \checkmark & 9 & 45 & May 11, 2015\\
        task\_behavior\_engine\footnote{\url{https://github.com/ToyotaResearchInstitute/task_behavior_engine}} & Python & & \checkmark & \checkmark & 9 & 27 & Aug 3, 2018\\
        
        gdxAI\footnote{\url{https://github.com/libgdx/gdx-ai}} & Java &  &  & \checkmark & 189 & 758 & Jul. 26, 2019\\
        bte2\footnote{\url{https://github.com/piotr-j/bte2}} & Java &  &  \checkmark & \checkmark & 5 & 27 & Aug 6, 2018\\

        Behavior3 \footnote{\url{https://github.com/behavior3}} & JavaScript & \checkmark &  & \checkmark & 135 & 335 & Oct. 21, 2018\\
        BehaviorTree.js\footnote{\url{https://github.com/Calamari/BehaviorTree.js}} & JavaScript &  &  & \checkmark & 29 & 207 & May 3, 2019\\
        
        NPBehave\footnote{\url{https://github.com/meniku/NPBehave}} & Unity3D/C\# & \checkmark &  & \checkmark & 52 & 306 & Oct. 24, 2019\\
        BT-Framework \footnote{\url{https://github.com/f15gdsy/BT-Framework}} & Unity3D/C\# &  &  & \checkmark & 66 & 163 & Mar 2, 2015\\
        fluid-behavior-tree \footnote{\url{https://github.com/ashblue/fluid-behavior-tree}} & Unity3D/C\# &  &  & \checkmark & 15 & 107 & Jun 13, 2019\\
        hivemind \footnote{\url{https://github.com/rev087/hivemind}} & Unity3D/C\# & \checkmark  &  & \checkmark & 13 & 47 & May 14, 2015\\
        
        BehaviorTree.CPP\footnote{\url{https://github.com/BehaviorTree/BehaviorTree.CPP}} & C++ & \checkmark & \checkmark & \checkmark & 77 & 306 & \emph{up to date}\\
        ROS-Behavior-Tree\footnote{\url{https://github.com/miccol/ROS-Behavior-Tree}} & C++ & \checkmark & \checkmark & \checkmark & 48 & 154 & Oct. 22, 2018\\
        Behavior Tree Starter Kit\footnote{\url{https://github.com/aigamedev/btsk}} & C++ &  &  & \checkmark & 103 & 297 & Jun 24, 2014\\
        BrainTree\footnote{\url{https://github.com/arvidsson/BrainTree}} & C++ &  &  & \checkmark & 27 & 121 & Aug 23, 2018\\
        Behavior-Tree\footnote{\url{https://github.com/miccol/Behavior-Tree}} & C++ & \checkmark &  & \checkmark & 33 & 116 & Oct 17, 2018\\
        
    \hline
    \end{tabular}
\end{table*}

\texttt{py\_trees} (extended to \texttt{py\_trees\_ros}\footnote{\url{https://github.com/splintered-reality/py_trees_ros}}) is one of the most used BT libraries, possibly because it is open source and well maintained. The core library is aimed at game AI design, but it has extensions for robotic implementations in ROS.
It features the following design constraints\footnote{from \url{https://py-trees.readthedocs.io/en/devel/background.html}}:
\begin{itemize}
\itemsep=0pt
    \item no interaction or sharing of data between tree instances;
    \item no parallelisation of tree execution;
    \item only one behaviour initialising or executing at a time.
\end{itemize}
Visualizing the BT is also possible thanks to {ASCII} or dot graph output. Information sharing between nodes is enabled by a \emph{blackboard}, which allow for sharing information within nodes in the same BT but not between different BTs.
A notable limitation of this library is that the \emph{Sequence} node is implemented \say{with memory}, thus removing one of the advantages of using a BT, i.e. \emph{reactivity}. \citep{colledanchise_behavior_2018} notes that \say{the use of nodes with memory is advised exclusively for those cases where there is no unexpected event that will undo the execution of the subtree in a composition with memory}. To overcome this limitation, each user needs to implement a memoryless version of the Sequence node, which is quite straightforward using Fallbacks and negations. The \emph{modularity} of BTs allows to program a specific sub-tree and then take it out-of-the-shelf when needed; however, the use of the \emph{blackboard} is inherently task specific, thus limiting \emph{modularity}.

\texttt{Owyl} is an early BT library and one of the first implemented in Python. To our knowledge it hasn't been widely used nor is still maintained. 

\texttt{gdxAI} is a framework written in Java for game development with \texttt{libGDX}. It supports features for AI agents movement, pathfinding and decision making, both via BTs and FSMs.

\texttt{Behavior3} is a library developed by the authors of \citep{pereira_framework_2015}. It features a visual editor, namely \texttt{Behavior3 Editor} (corresponding to the data in Table \ref{tab:lib}), which can export the modeled trees to JSON files, and to a client library for Python and Javascript. Even though a visual editor simplifies the design and the readability, an extensive documentation, as well as periodic maintenance are lacking. ROS support is also not implemented.

\texttt{BehaviorTree.js} is another library addressed to the design of game AI implemented in Javascript. Note that the return state of the nodes of this library can be only \emph{Success} or \emph{Failure}, no \emph{Running}.

\texttt{NPBehave} is a BT library that targets game AI for the Unreal Engine as are \texttt{BT-Framework},
\texttt{fluid}-\texttt{behavior}-\texttt{tree} and \texttt{hivemind}, which comes with a visual editor with runtime visual debugging capabilities. In particular, \texttt{NPBehave} is an event driven library, i.e., it features event driven behavior trees which don't need to be  executed from the root node at each tick.
This design is claimed to be more efficient and simpler to use.

\texttt{ROS-Behavior-Tree} is a library implemented by the authors in~\citep{colledanchise_how_2014}. According to the documentation, the latest version of ROS is not currently supported and the last commit is dated October 2018. A probable reason for this state is because the authors collaborated in the creation of \texttt{BehaviorTree.CPP} which features the Groot GUI Editor and an implementation in ROS.

\section{Open challenges}
\label{sec:end}
In this section we point out four important open challenges for BT research: \emph{Explainable AI, Human-robot-inter\-action, safe AI}, and \emph{the combination of learning and BTs}.

Explainable AI is an area of increasing importance, especially in the light of the recent success of data driven methods such as Reinforcement Learning and Deep Learning.
BTs are widely believed to provide good explainability, as they have been used to capture human workflows~\citep{deneke_conceptual_2017,hannaford_behavior_2018}, and are in general well suited for manual design. Furthermore, Hierarchical Task Networks (HTNs) is a planning tool built on the intuitive idea of high level tasks being composed of low level tasks in several layers, and the parallels between BTs and HTNs have been explored in~\citep{neufeld_hybrid_2018}. Finally, human readability is a well known advantage of BTs~\citep{jones_evolving_2018,paxton_evaluating_2018,shu_behavior_2019}. Together, these ideas indicate that BTs can have a role to play in explainable AI, either by incorporating data driven methods into BTs, or by learning a BT from the workings of a data driven policy.

Human-robot-interaction (HRI) and collaboration is another area that will increase in importance as robots are more integrated in factories and homes. Explainable AI is certainly a part of HRI, but areas such as programming by non experts, and human in the loop collaborations are also important. Again, the human readability of BTs, together with \emph{modularity}, will simplify these issues.  Important progress has already been made in~\citep{paxton_costar_2017,paxton_user_2017,paxton_evaluating_2018}, but much remains to be done.

Safe autonomy is a third area related to explainable AI and HRI. To be able to guarantee safety of a robot system will be increasingly important as autonomous systems that are big enough to harm humans, such as unmanned vehicles, start to share spaces with people. 
The \emph{transparency} and \emph{modularity} of BTs can play an important role in this area. Some initial work in this area was done in~\citep{sprague_adding_2018}, but many questions remain.

Finally, as seen in Section~\ref{sec:learning}, the combination of BTs and ML is being explored by several research groups. It is clear that end-to-end RL will not provide a viable solution to many problems, as in some cases the state space is just to large to explore efficiently, and in others, lack of explainability and/or safety guarantees makes such approaches infeasible. However, both these problems could be addressed by a BT solution where individual nodes, leaves as well as interior ones, are improved by learning methods. 

\section{Conclusions}
\label{sec:conclusions}
In this survey paper we have provided an overview of the over 160 research papers devoted to the development of BT as a tool for AI in games and robotics. We have partitioned and analyzed the papers based on both application areas and methods used, and finally provided a description of a number of open challenges to the research community.

\section*{Acknowledgments}

This work was partially supported by the Swedish Foundation for Strategic Research, Vinnova NFFP7 (grant no. 2017-04875) and by the Wallenberg AI, Autonomous Systems, and Software Program (WASP) funded by the Knut and Alice Wallenberg Foundation. The authors gratefully acknowledge this support.

\bibliographystyle{apalike}
\bibliography{biblioShort,non_bt_papers}

\end{document}